\documentclass[10pt,twocolumn,letterpaper]{article}

\usepackage{iccv}              

\definecolor{iccvblue}{rgb}{0.21,0.49,0.74}
\usepackage[pagebackref,breaklinks,colorlinks,allcolors=iccvblue]{hyperref}
\usepackage{algorithm}
\usepackage{algpseudocodex}
\usepackage{amsmath}
\usepackage{amssymb}
\usepackage{multirow}
\usepackage{graphicx}
\usepackage{booktabs}
\usepackage{makecell}
\usepackage{subcaption}
\usepackage{fancybox}
\usepackage{fancyvrb}
\usepackage{bm}
\usepackage{multirow}
\usepackage{booktabs}
\usepackage{bbding}
\usepackage{makecell}
\usepackage{subcaption}
\usepackage{fancybox}
\usepackage{fancyvrb}
\usepackage{colortbl}
\usepackage{xcolor}
\usepackage{listings}

\definecolor{ourblue}{rgb}{0.0, 0.0, 1.0}
\newcommand{\dd}[2]{$#1\scriptstyle{\pm#2}$}
\newcommand{\ddbf}[2]{$\mathbf{#1\scriptstyle{\pm#2}}$}
\definecolor{softred}{RGB}{255, 178, 178}  
\definecolor{softorange}{RGB}{255, 218, 179} 
\definecolor{softyellow}{RGB}{255, 244, 191} 
\def\paperID{9479} 
\def\confName{ICCV}
\def\confYear{2025}

\title{Spatial-Temporal Aware Visuomotor Diffusion Policy Learning}

\author{Zhenyang Liu$^{1,2}$, Yikai Wang$^{3}$, Kuanning Wang$^{1}$, Longfei Liang$^{4}$\\ Xiangyang Xue$^{1}$\footnotemark[2], Yanwei Fu$^{1,2}$\footnotemark[2]\\
$^1$\normalsize Fudan University
$^2$\normalsize Shanghai Innovation Institute
$^3$\normalsize Nanyang Technological University $^4$\normalsize NeuHelium Co., Ltd\\
{\tt\small liuzy24@m.fudan.edu.cn, yikai.wang@ntu.edu.sg, knwang24@m.fudan.edu.cn} \\
{\tt\small longfei.liang@neuhelium.com,
\{xyxue,yanweifu\}@fudan.edu.cn}\\
Project Page: \href{https://zhenyangliu.github.io/DP4/}{ZhenyangLiu.github.io/DP4}}

\begin{document}
\maketitle
\renewcommand{\thefootnote}{\fnsymbol{footnote}}
\footnotetext[2]{Corresponding authors.}
\footnotetext[3]{Prof. Yanwei Fu is also with Institute of Trustworthy Embodied Al, and the School of Data Science, Fudan University.}
\begin{abstract}
Visual imitation learning is effective for robots to learn versatile tasks. However, many existing methods rely on behavior cloning with supervised historical trajectories, limiting their 3D spatial and 4D spatiotemporal awareness. Consequently, these methods struggle to capture the 3D structures and 4D spatiotemporal relationships necessary for real-world deployment. In this work, we propose 4D Diffusion Policy (DP4), a novel visual imitation learning method that incorporates spatiotemporal awareness into diffusion-based policies. Unlike traditional approaches that rely on trajectory cloning, DP4 leverages a dynamic Gaussian world model to guide the learning of 3D spatial and 4D spatiotemporal perceptions from interactive environments. Our method constructs the current 3D scene from a single-view RGB-D observation and predicts the future 3D scene, optimizing trajectory generation by explicitly modeling both spatial and temporal dependencies. Extensive experiments across 17 simulation tasks with 173 variants and 3 real-world robotic tasks demonstrate that the 4D Diffusion Policy (DP4) outperforms baseline methods, improving the average simulation task success rate by 16.4\% (Adroit), 14\% (DexArt), and 6.45\% (RLBench), and the average real-world robotic task success rate by 8.6\%.
\end{abstract}

\section{Introduction}
Visual Imitation learning has emerged as an effective method for training robots to perform complex tasks, such as object grasping~\cite{wang2023mimicplay,johns2021coarse,shridhar2023perceiver}, legged locomotion~\cite{peng2020learning,yang2023generalized}, dexterous manipulation~\cite{haldar2023teach,qin2022dexmv}, and mobile manipulation~\cite{du2022bayesian,shafiullah2023bringing}. Recent advances in computer vision have further enhanced visual imitation learning, enabling impressive performance across various applications~\cite{shafiullah2022behavior,shridhar2023perceiver,hansen2022pre,chen2021decision}. 
Progress in computer vision has enabled sophisticated visual imitation learning methods that demonstrate excellent performance across various applications~\cite{shafiullah2022behavior,shridhar2023perceiver,hansen2022pre,chen2021decision}. However, real-world robotic deployment demands robust generalization to unseen scenarios, necessitating a deeper understanding of the spatiotemporal physical rules that govern dynamic environments.

\begin{figure}[t]
	\centering
	\includegraphics[width=1\linewidth]{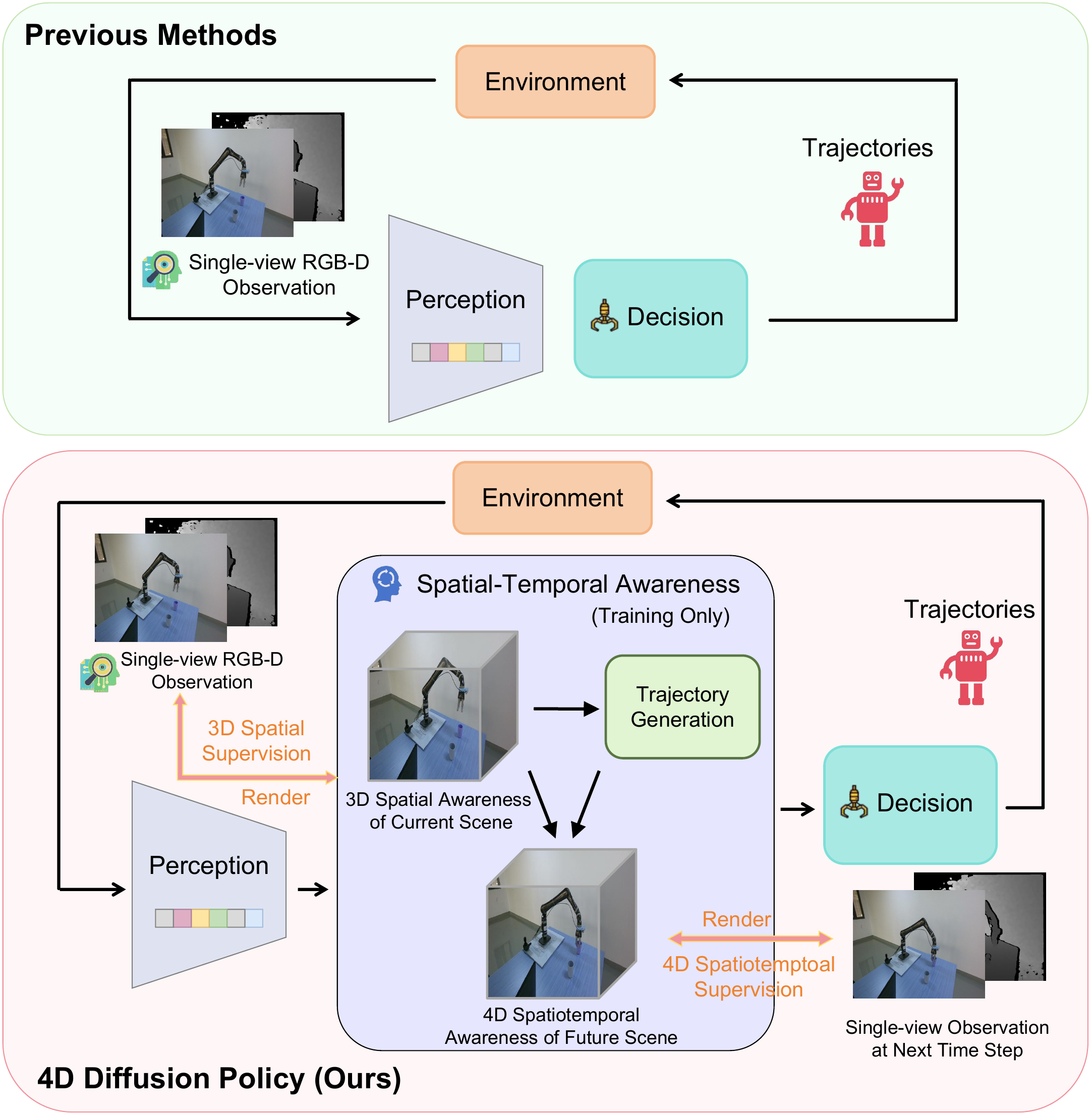}
    \vspace{-0.20in}
	\caption{\textbf{Spatial-temporal awareness in the 4D Diffusion Policy (DP4).}\label{teaser} 
    Previous methods train perception and decision-making with trajectory supervision, but trajectory cloning fails to capture the 3D spatial and 4D spatiotemporal relationships. In contrast, DP4 constructs the current 3D scene with 3D spatial supervision from a single RGB-D view and predicts future 3D scene candidates using 4D spatiotemporal supervision, optimizing trajectory generation by effectively capturing both 3D structures and 4D dependencies.
    }
    \vspace{-0.25in}
\end{figure}

Despite their success, existing visual imitation learning methods struggle to capture the 3D structure of environments and 4D spatiotemporal dependencies of interactive tasks. These limitations arise mainly because current approaches rely on behavior cloning from \textit{static, supervised trajectories}. Such methods often fail to infer the essential physical properties needed for spatial reasoning, object interaction, and environment-aware decision-making.

In this work, we propose 4D Diffusion Policy (DP4), a novel visual imitation learning method that integrates spatial-temporal awareness into diffusion-based policies. Unlike prior works that passively mimic demonstrations exclusively from supervised trajectories, DP4 actively learns from a dynamic Gaussian world model~\cite{kerbl20233d, wu20244d}, which explicitly encodes 3D spatial structures and 4D spatiotemporal variations as part of the imitation learning process. Thus,
as illustrated in Fig.~\ref{teaser}, DP4 extends conventional imitation learning pipelines in the following ways:
\begin{itemize}
    \item \textit{3D spatial perception is learned via Gaussian Splatting (3DGS)~\cite{kerbl20233d} }, ensuring accurate reconstruction of the current scene's geometry from point cloud inputs.
    \item \textit{4D spatiotemporal awareness is enforced by future 3DGS reconstruction}, where the system learns to predict and adaptively refine future environmental states based on the agent's planned trajectories.
    \item  \textit{Trajectory generation is conditioned on these structured spatial-temporal representations}, allowing the diffusion policy model to reason over future states rather than just replicating past behaviors.
    
\end{itemize}

So our method first converts RGB-D observations into 3D point clouds, from which we extract global and local spatiotemporal representations. These serve as conditions for the diffusion policy model to guide next-step trajectory generation. To strengthen structured perception:

\begin{itemize}
    \item \textit{We present a novel Gaussian regressor} that learns to reconstruct the 3D scene from point cloud inputs, ensuring spatially consistent representations.
    \item \textit{We enforce temporal consistency} by deforming the current scene's 3DGS into the predicted future state, supervising this process using ground-truth RGB-D reconstructions from the future scene.
\end{itemize}

\begin{figure*}
	\centering
	\includegraphics[width=0.9\linewidth]{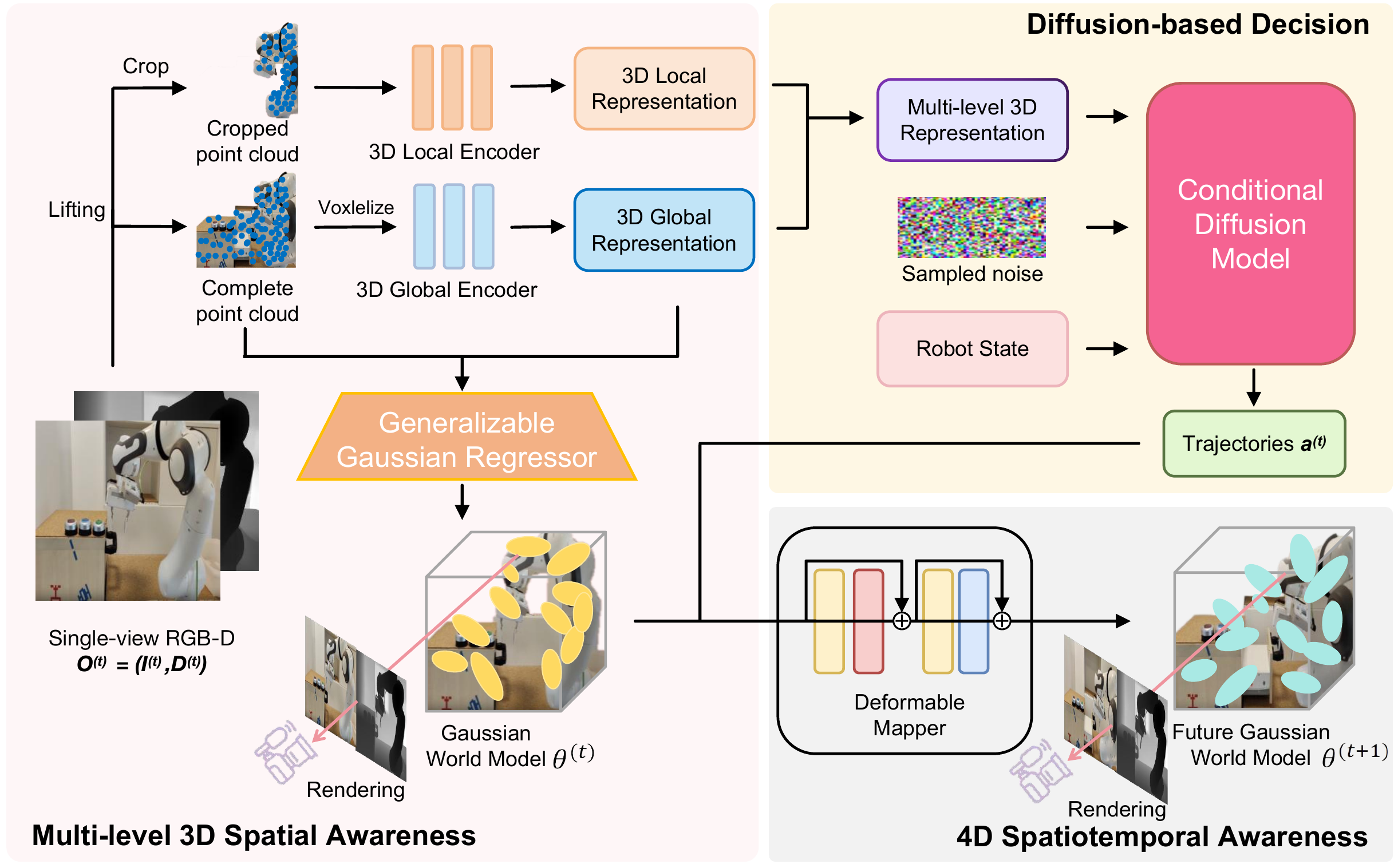}
    \vspace{-0.10in}
\caption{\textbf{
The framework of our 4D Diffusion Policy (DP4).} From a single-view RGB-D observation, we construct 3D point clouds and extract global and local features to enrich both holistic and focused perceptions. These multi-level representations condition the diffusion policy model to generate trajectories based on current robot states. We introduce a Gaussian world model in DP4 to capture 3D structures and 4D spatiotemporal relationships. The current observation’s 3D Gaussian Splatting (3DGS) is derived via a generalizable Gaussian regressor from point clouds and multi-level features. By enforcing consistency between ground-truth and rendered RGB-D images from this 3DGS, we enhance 3D spatial awareness. Additionally, future 3DGS are predicted from current states using policy-generated trajectories, with rendered RGB-D consistency fostering 4D spatiotemporal awareness. This improved spatiotemporal representation significantly benefits complex tasks such as object grasping and dexterous manipulation.
} \label{fig:pipelne}
    \vspace{-0.05in}
\end{figure*}

We evaluate DP4 extensively across 17 simulated tasks (173 variants) from \textbf{four benchmarks, along with three real-world tasks}, including the challenging task of \textbf{dexterous manipulation of deformable objects}. Our experiments show that DP4 significantly outperforms state-of-the-art multi-task visual imitation methods, achieving a +16.4\% success rate on Adroit, +14\% on DexArt, +6.45\% on RLBench, and +8.6\% improvement in real-world robotic tasks.

Our work bridges the gap between visual imitation learning and spatial-temporal physical reasoning, demonstrating the importance of structured world models in improving generalization and robustness in real-world robotic applications, with these contributions:
\begin{itemize}

\item \textbf{\textit{Spatiotemporal-centric visual imitation learning}}: Unlike existing methods that overlook 3D perception and dynamic interactions, DP4 explicitly models 3D spatial structures and 4D temporal dynamics, improving generalization in dynamic environments.
\item \textbf{\textit{4D Diffusion Policy with structured awareness}}: We introduce a diffusion-based visual imitation learning framework that generates trajectories based on learned spatial-temporal world representations.
\item \textbf{\textit{Dynamic Gaussian World Model for structured supervision}}: Our model actively learns from real-world interactions, ensuring the embedding of 3D spatial and 4D temporal reasoning into policy learning.
\item \textbf{\textit{State-of-the-art performance across diverse tasks}}: DP4 sets unprecedented success rates across 17 simulated and real-world robotic tasks, setting a new benchmark for imitation learning in dynamic environments.
\end{itemize}

\section{Related Work}
\noindent\textbf{Visual imitation learning} \cite{kim24openvla,liu2024rdt,li2025gr} is an effective method for equipping robots with skills, typically relying on numerous expert demonstrations as observation-action pairs. Recent advanced visual imitation learning methods commonly employ 2D images\cite{florence2022implicit,mandlekar2021matters,arunachalam2023dexterous,shafiullah2022behavior,wang2023mimicplay,chi2023diffusion} as state representations for their rich global information and ease of acquisition from raw sensory inputs, whereas 3D representations (e.g., point clouds and object geometries)~\cite{zhu2025point,Ze2024DP3} are used for their explicit spatial structure, enhancing the spatial reasoning. However, these methods overlook the spatiotemporal dynamics of scenes arising from behavior-environment interactions. DP4 overcomes this limitation by employing a dynamic Gaussian world model to derive scene-level 3D structures from current scene reconstruction and 4D spatiotemporal dynamics from future scene reconstruction.

\noindent\textbf{Diffusion models as policies}\quad
Building on their success in image and video generation~\cite{yang2023diffusion,chen2023diffusiondet,ho2022video}, diffusion models have emerged as effective policy frameworks in robotics~\cite{qin2022dexmv,mao2024diffusion,yang2023policy,chi2023diffusion,Ze2024DP3}. They are primarily applied in reinforcement learning and visual imitation learning, with Diffusion Policy~\cite{chi2023diffusion} and variants~\cite{wang2022diffusion,Ze2024DP3} achieving state-of-the-art results by fusing visual inputs (e.g., 2D images, 3D point clouds) with complex, multi-modal action spaces. Nevertheless, these approaches often underutilize visual perception, relying heavily on behavior cloning from supervised trajectories. In contrast, DP4 introduces a Gaussian world model that reconstructs current scenes from RGB-D data via volumetric rendering for 3D spatial understanding, and employs a deformation Gaussian field to predict future scene distributions through environment interaction, enabling comprehensive 4D spatiotemporal perception.

\noindent\textbf{World model}\quad 
Recent advances demonstrate that world models effectively encode scene dynamics by predicting future states from current states and actions, with applications in autonomous driving~\cite{hu2022model,hu2023gaia,wang2024drivedreamer} and robotic manipulation~\cite{seo2023masked,wu2023daydreamer}. Early works~\cite{ha2018recurrent,hafner2022deep,hu2022model,hafner2019dream,hafner2020mastering} employed latent space autoencoding for future prediction, yielding strong results in both simulated and real environments. Nonetheless, learning accurate latent representations is data-intensive and generally confined to simple tasks due to limited implicit feature capacity. To address these challenges, this study integrates Gaussian Splatting~\cite{kerbl20233d}, known for its efficiency and rendering quality, into world model construction. The resulting 4D Diffusion Policy leverages a dynamic Gaussian world model to capture scene-level 3D structure and 4D spatiotemporal dynamics.

\section{Methodology}
In this section, we introduce 4D Diffusion Policy (DP4), a novel visual imitation learning method that emphasizes spatial-temporal awareness in diffusion policies. DP4 employs 3D Gaussian Splatting for informative supervision, enabling the learning of 3D spatial and 4D spatiotemporal perceptions in interactive environments. The DP4 framework is illustrated in Figure~\ref{fig:pipelne}.

\subsection{Problem Setup}
Visual imitation learning allows robots to efficiently develop human-like skills using observation-action pairs from expert demonstrations. The intelligent agent must interactively predict subsequent continuous trajectories based on the current observation. The visual observation at time step $t$, denoted $\mathbf{O}^{(t)}=(\mathbf{C}^{(t)}, \mathbf{D}^{(t)})$, is derived from expert demonstrations, where $\mathbf{C}^{(t)}$ and $\mathbf{D}^{(t)}$ represent single-view RGB and depth images, respectively. Using the visual observation $\mathbf{O}^{(t)}$, the visuomotor policy learning model predicts the optimal action $\mathbf{a}^{(t)}$ at current time step and generates subsequent optimal continuous action sequence {$\mathbf{a}^{(t+1)}$}.

To effectively train a visuomotor policy learning model, expert demonstrations are supplied as offline datasets for visual imitation learning, consisting of sample triplets with visual observations and expert actions~\cite{wang2022diffusion,mandlekar2021matters}. Current methods utilize robust visual representations, such as Diffusion Policy and 3D Diffusion Policy, to derive informative latent features for optimal action prediction. However, these methods heavily depend on behavior cloning with supervised historical trajectories. This approach struggles to fully capture 3D structures and often overlooks 4D spatiotemporal dynamics. Consequently, the predicted actions frequently fail to achieve complex human goals, particularly when accurate object interactions are absent. In this study, we introduce the 4D Diffusion Policy, which employs a dynamic Gaussian world model to extract 3D structures from current scene reconstruction and 4D spatiotemporal dynamics from future scene reconstruction.

\subsection{Multi-level 3D Spatial Awareness}\label{3.2}
To capture complex 3D structures in the interactive physical world, DP4 achieves multi-level spatial perception by generating 3D representations from local point clouds and global voxel grids. These 3D representations are derived from an RGB-D image captured by a single-view camera with known extrinsic and intrinsic parameters, enhancing practical applicability in real-world settings. DP4 avoids impractical setups, such as using multiple cameras during training and inference or requiring extensive prior environmental visual data. In both simulated and real-world settings, we acquire depth images from a single camera and convert them into complete point clouds using the camera’s extrinsic and intrinsic parameters.

\noindent\textbf{Learning 3D representations}\quad To generate the 3D local representation, DP4 crops the point clouds, retaining only points within a specified bounding box. This allows DP4 to focus on critical scene elements (e.g., the robot arm) and exclude redundant points (e.g., the ground). DP4 encodes the cropped point clouds into compact 3D local representations using a lightweight MLP network, called the 3D Local Encoder. To obtain the 3D global representation, DP4 converts the point clouds at step $t$ into a $100^3$ voxel. The 3D Global Encoder encodes this voxel, outputting the 3D global representation $v^{(t)} \in \mathbb{R}^{100^3 \times 64}$.

\noindent\textbf{Enhance 3D spatial awareness}\quad To enhance the representation $v^{(t)}$ with global structure and texture information, we employ a Generalizable Gaussian Regressor, which takes the deep volume $v$ as the scene representation to derive the Gaussian world model based on Gaussian Splatting~\cite{kerbl20233d}.
The Generalizable Gaussian Regressor is learned by rendering RGB and depth images from the constructed Gaussian world model. The entire neural rendering process is described below. Define $v^{(t)}\mathbf{x} \in \mathbb{R}^{128}$ as the sampled 3D feature at point $\mathbf{x}$ from the 3D global representation $v^{(t)}$. $v^{(t)}\mathbf{x}$ is obtained through trilinear interpolation, given the discretized nature of the 3D global representation $v^{(t)}$. The Generalizable Gaussian Regressor uses the 3D coordinates $\mathbf{x}$ and the feature $v^{(t)}_\mathbf{x}$ to parameterize the Gaussian primitive as $\theta^{(t)} = (\mu^{(t)}, c^{(t)}, r^{(t)}, s^{(t)}, \sigma^{(t)})$, representing the position, color, rotation, scale, and opacity of the Gaussian primitive. To render a single view, we project Gaussian primitives onto the 2D plane using differential tile-based rasterization.	The pixel value $\mathbf{p}$ is rendered using alpha-blend rendering:	
\begin{equation}\label{eq:rgb_depth_rendering}
\begin{aligned}
C^{(t)} (\mathbf{p}) &=\sum_{i=1}^{N} \alpha_i^{(t)}  c_i^{(t)}  \prod_{j=1}^{i-1} (1-\alpha_j^{(t)} ),\\
D^{(t)} (\mathbf{p}) &= \sum_{i=1}^{N} \alpha_i^{(t)}  d_i^{(t)}  \prod_{j=1}^{i-1} (1-\alpha_j^{(t)} ),
	\end{aligned}
\end{equation}
where $\alpha_i^{(t)} = \sigma_i^{(t)} \exp{-\frac{1}{2} (\mathbf{p} - \mu_i^{(t)})^{\top} \Sigma_i^{-1} (\mathbf{p} - \mu_i^{(t)})}$ defines the 2D density of Gaussian points during the splatting process, $C^{(t)}$ and $D^{(t)}$ represent the rendered RGB image and depth, respectively, $N$ is the number of Gaussians per tile, $d_i^{(t)}$ is the $z$-value obtained by transforming coordinate $u$ from world to camera coordinates, and $\Sigma_i$ denotes the covariance matrix derived from the Gaussian parameters’ rotation and scale. Our Generalizable Gaussian Regressor and 3D representation are optimized to reconstruct RGB and depth images from a single view by minimizing a 3D loss function:
\begin{equation}
\mathcal{L}_{\text{3D}}=\sum_{\mathbf{p} }\|\mathbf{C}^{(t)} (\mathbf{p})-\mathbf{C}^{*(t)}\|_2^2 +  \|\mathbf{D}^{(t)} (\mathbf{p})-\mathbf{D}^{*(t)}\|_2^2,
\end{equation}
where $\mathbf{C}^{*(t)}$ and $\mathbf{D}^{*(t)}$ represent the ground truth color and depth at time step $t$.	This 3D loss allows the 3D representation to emphasize local visual features while incorporating global structural knowledge and texture information.	

\subsection{4D Spatiotemporal Awareness}
To emphasize 4D spatiotemporal awareness in the diffusion policy, DP4 integrates dynamics into the Gaussian world model described in Section \ref{3.2}. This model parameterizes the Gaussian mixture distribution for dynamic Gaussian splatting, enabling future scene reconstruction via parameter propagation. Thus, the model gains valuable supervision in interactive environments by ensuring consistency between reconstructed and real feature scenes. Specifically, the model learns environmental dynamics by predicting the future state $s^{(t+1)}$ from the current state $s^{(t)}$ and action $a^{(t)}$ at time step $t$.

In the dynamic Gaussian world model, we define the current state as the visual observation at the current time step, with actions representing the robot components. These observations and actions predict visual scenes for the next time step, representing the future state. Building on the Gaussian world model, DP4 uses a deformable MLP $p_{\phi}$ to estimate changes in Gaussian parameters during time propagation and a Gaussian renderer $\mathcal{R}$ to generate RGB and depth images for the future state:
\begin{equation}\label{eq:deform}
	\begin{aligned}
&\Delta \theta^{(t)}= p_\phi\left(\theta^{(t)}, a^{(t)}\right), \\
&\theta^{(t+1)} = \theta^{(t)} + \Delta \theta^{(t)}, \\
&C^{(t+1)} (\mathbf{p}), D^{(t+1)} (\mathbf{p}) = \mathcal{R}\left(\theta^{(t+1)}, \mathbf{p}\right),
	\end{aligned}
\end{equation}
where ${\mathbf{C}}^{(t+1)}(\mathbf{p})$ and ${\mathbf{D}}^{(t+1)}(\mathbf{p})$ represent the predicted RGB and depth images at time step $t+1$, $\theta^{(t)}$ represents the 3D Gaussian primitive parameters at time step $t$, $w$ denotes the camera pose used to project the Gaussian primitives, and $\mathcal{R}$ is the rendering process from Equation \ref{eq:rgb_depth_rendering}. DP4 employs multi-head neural networks as the Gaussian regressor, with each head predicting distinct features of the Gaussian parameters. By estimating changes in position and rotation between time steps, we derive the Gaussian parameters for the future step. The Gaussian renderer then projects the propagated Gaussian distribution onto the single view to reconstruct the future scene. To model 4D spatiotemporal relationships in the physical world, we enforce consistency supervision between reconstructed and real scenes at time step $t+1$, embedding scene-level spatiotemporal dynamics into the Gaussian parameters. Specifically, the training objective aligns predicted future scenes, derived from current observations and actions, with real scenes, as formulated below:
\begin{equation}\label{ref}
\small
\mathcal{L}_{\text{4D}}=\sum_{\mathbf{p} }\|\mathbf{C}^{(t+1)} (\mathbf{p})-\mathbf{C}^{*(t+1)}\|_2^2 +  \|\mathbf{D}^{(t+1)} (\mathbf{p})-\mathbf{D}^{*(t+1)}\|_2^2,
\end{equation}
where $\mathbf{C}^{(t+1)}$ and $\mathbf{D}^{(t+1)}$ represent the real future counterparts. This additional task of predicting future scenes from current observations via the dynamic Gaussian world model serves as a spatiotemporal enhancement penalty. Training with this task, the multi-level 3D representation learns to encode physical properties and develop awareness of 4D spatiotemporal dynamics.

\subsection{Diffusion-based Decision}
The multi-level 3D representation, including local and global 3D representations, is optimized to predict the robot's optimal actions. We model this process as a conditional denoising diffusion framework, conditioned on the multi-level 3D representation $r$ and the robot state $q$, which converts randomly sampled Gaussian noise into actions $a$. Starting with sampled Gaussian noise $a^K$, the denoising network $\boldsymbol{\epsilon}_\theta$ performs $K$ iterations to progressively convert it into the noise-free action $a^0$.
\begin{equation}
{a}^{k-1}=\alpha_k\left(a^k-\gamma_k \boldsymbol{\epsilon}_\theta\left(a^k, k, r, q\right)\right)+\sigma_k \mathcal{N}(0, \mathbf{I}),
\end{equation}
where $\mathcal{N}(0, \mathbf{I})$ represents the sampled Gaussian noise, and $\alpha_k$, $\gamma_k$, and $\sigma_k$ are functions of $k$ defined by the noise scheduler. To train the denoising network $\boldsymbol{\epsilon}_\theta$, we randomly sample a data point $a^0$ from the dataset and use a diffusion process to generate the noise $\boldsymbol{\epsilon}^k$ at iteration $k$. The training objective is to predict the noise added to the data.
\begin{equation}
\mathcal{L}_{\text{action}}=\text{MSE}\left(\boldsymbol{\epsilon}^k, \boldsymbol{\epsilon}_\theta(\bar{\alpha_k} a^0 + \bar{\beta_k} \boldsymbol{\epsilon}^k, k, r, q)\right)\,,
\end{equation}
where $\bar{\alpha_k}$ and $\bar{\beta_k}$ are noise schedule that performs one step noise adding. 

The overall learning objective for DP4 is as follows:
\begin{equation}
    \mathcal{L}_\text{DP4} = \mathcal{L}_{\text{action}} + 
\lambda_{\text{3D}} \mathcal{L}_{\text{3D}} + \lambda_{\text{4D}} \mathcal{L}_{\text{4D}}\,,
\end{equation}
where $\lambda_{\text{3D}}$ and $\lambda_{\text{4D}}$ balance the 3D loss for current scenes and the 4D loss for future scenes across different objectives. We train DP4 using a joint training strategy. Empirical observations show that this strategy enhances information fusion while learning shared features. During training, we introduce a warm-up phase, freezing the deformable mapper for the first 500 iterations to establish a stable multi-level 3D compact representation and Gaussian regressor. After the warm-up phase, we simultaneously train the entire Gaussian world model and the diffusion-based decision module. After optimization, DP4 generates optimal actions with spatiotemporal awareness, including both 3D spatial and 4D spatiotemporal perceptions.

\section{Experiments}
\subsection{Experiment Setup}
\noindent\textbf{Simulations}\quad
For simulation experiments, we gathered 22 tasks across various domains, covering a wide range of robotic skills. These tasks span challenging scenarios, such as bimanual, deformable object, and articulated object manipulation, as well as simpler tasks like parallel gripper manipulation. The tasks include Adroit tasks~\cite{rajeswaran2017learning} (Door, Pen, Hammer), DexArt tasks~\cite{bao2023dexart} (Laptop, Faucet, Bucket, Toilet), and RLBench tasks~\cite{james2020rlbench} (10 challenging tasks, each with at least two variations, totaling 166).

\noindent\textbf{Real robot}\quad 
We used a KINOVA GEN2 robot with a Realsense D455 depth camera mounted in an eye-to-hand configuration.
In an indoor environment, we arranged several bottles and cups to enable the agent to generalize manipulation skills across scenes, designing three tasks, including \textit{Grasping bottles}, \textit{Pouring the water from the bottle into the cup} and \textit{stacking cups}. The RealSense D455 camera records the entire scene and robot state from a third-person perspective. This setup supplies RGB-D observations for policy training and RGB and depth supervision for constructing the dynamic Gaussian world model.

\noindent\textbf{Expert demonstration}\quad
Expert trajectories for ManiSkill and Adroit are collected using RL-trained agents, using VRL3~\cite{wang2022vrl3} for Adroit and PPO~\cite{schulman2017proximal} for other domains. We generate successful trajectories with RL agents and ensure that all visual imitation learning algorithms use identical demonstrations. Additionally, we collect 20 demonstrations for each RLBench task using the motion planner.

\noindent\textbf{Baselines}\quad 
The main objective of DP4 is to highlight the importance of 3D spatial and 4D spatiotemporal awareness in diffusion policies.	 To this end, our primary baseline is the point cloud-based 3D Diffusion Policy (DP3)~\cite{Ze2024DP3}.	 Additionally,
we incorporate comparisons with other visual imitation learning methods: IBC~\cite{florence2022implicit}, BCRNN~\cite{mandlekar2021matters}, Diffusion Policy (DP)~\cite{chi2023diffusion}, PerAct~\cite{shridhar2023perceiver}, GNFactor~\cite{ze2023gnfactor}, and ManiGaussian~\cite{lu2024manigaussian}. 
Note that image and depth resolutions for all 2D and 3D methods remain consistent across experiments, ensuring fair comparisons.	

\noindent\textbf{Evaluation metric}\quad
Each experiment is conducted using three seeds. For Adroit and DexArt simulation tasks, we assess 20 episodes every 200 epochs and calculate the mean of the top five success rates. For RLBench tasks, we evaluate 25 episodes per task at the final checkpoint for 10 challenging tasks. We report the mean and standard deviation of success rates across the three seeds.

\noindent\textbf{Implementation details}\quad
We implement a convolutional network-based diffusion policy using DDIM\cite{song2020denoising} as the noise scheduler, preferring sample prediction to enhance high-dimensional action generation. Training employs 100 timesteps, while inference uses 10. Models are trained for 3,000 epochs with a batch size of 32 on both simulated and real-world tasks. Loss hyperparameters $\lambda_{\text{3D}}$ and $\lambda_{\text{4D}}$ are set to 0.1 and 0.01, respectively (see Section \ref{sec.act} for details).
For fair comparison with DP3\cite{Ze2024DP3}, we adopt the same 3D Local Encoder: a three-layer MLP, max-pooling for order-equivariant aggregation, and a projection head generating compact 3D local features, with LayerNorm inserted for training stability.
All experiments run on an NVIDIA H100 80GB GPU and a 192-vCPU Intel Xeon Platinum 8468.

\subsection{Quantitative Comparison}
We compare our DP4 with previous state-of-the-art methods on the Adroit, DexArt, and RLBench simulation tasks.

\noindent \textbf{Adroit simulation}\quad
Table \ref{table:sim_adriot} presents the success rates for the Adroit simulation tasks.
DP4 achieves an overall success rate of 84.7\%, significantly outperforming 2D-based methods, including IBC~\cite{florence2022implicit}, BCRNN~\cite{mandlekar2021matters}, and Diffusion Policy~\cite{chi2023diffusion}, while surpassing 3D-based methods such as 3D Diffusion Policy~\cite{Ze2024DP3}. Experimental results demonstrate that DP4 completes complex tasks more accurately by leveraging spatiotemporal perception.

\noindent \textbf{DexArt simulation}\quad
Table \ref{table:sim_dexart} presents the success rates for various DexArt simulation tasks. The DP4 achieves a success rate of 82.5\%, significantly outperforming 2D-based methods (e.g., IBC, BCRNN, and Diffusion Policy) by 33.5\% and 3D-based methods (e.g., 3D Diffusion Policy) by 14.0\%. Experimental results demonstrate that DP4 integrates 3D spatial and 4D spatiotemporal perception of the physical world with diffusion policies, achieving higher success rates across multiple tasks.

\begin{table}[t] \small
	\setlength{\tabcolsep}{8.6pt} 
	\centering
	\small 
	\begin{tabular}{lcccc}
		\toprule
		\textbf{Method} & \textit{Hammer} & \textit{Door} & \textit{Pen} & \textbf{Overall}  \\
		\midrule
		IBC~\cite{florence2022implicit}& \dd{0}{0}& \dd{0}{0} & \dd{9}{2} & 3.0\\
		BCRNN~\cite{mandlekar2021matters}& \dd{0}{0} & \dd{0}{0} & \dd{9}{3} & 3.0\\
		DP~\cite{chi2023diffusion} & \cellcolor{softyellow}\dd{48}{17} & \cellcolor{softyellow}\dd{50}{5} & \cellcolor{softyellow}\dd{25}{4} & \cellcolor{softyellow} 41.0\\
		DP3~\cite{Ze2024DP3} & \cellcolor{softorange}\dd{100}{0} & \cellcolor{softorange}\dd{62}{4} & \cellcolor{softorange}\dd{43}{6} & \cellcolor{softorange} 68.3\\
		\textbf{DP4 (ours)} & \cellcolor{softred}\ddbf{100}{0} & \cellcolor{softred}\ddbf{80}{2} & \cellcolor{softred}\ddbf{75}{3} & \cellcolor{softred}\textbf{84.7} \\
		\bottomrule
	\end{tabular}
        \vspace{-0.10in}
	\caption{\textbf{Success rates (\%) on the Adroit simulation.} We compare our \textbf{DP4} with more baselines in the Adroit simulation. The baselines include Implicit Behavioral Cloning (IBC), BCRNN, Diffusion Policy (DP), and 3D Diffusion Policy (DP3).}
        \vspace{-0.05in}
	\label{table:sim_adriot}
\end{table}

\begin{table}[t] \small
	\setlength{\tabcolsep}{5.2pt} 
	\centering
	\small 
	\begin{tabular}{lccccc}
		\toprule
		\textbf{Method} & \textit{Laptop} & \textit{Faucet} & \textit{Bucket} & \textit{Toilet} & \textbf{Overall}  \\
		\midrule
		IBC~\cite{florence2022implicit}& \dd{3}{2} & \dd{7}{1} & \dd{14}{1} & \dd{0}{0} & 6.0\\
		BCRNN~\cite{mandlekar2021matters}& \dd{3}{3} & \dd{1}{0} & \dd{5}{5} & \dd{0}{0} & 2.3\\
		DP~\cite{chi2023diffusion} & \cellcolor{softyellow}\dd{69}{4} & \cellcolor{softyellow}\dd{23}{8} & \cellcolor{softyellow}\dd{58}{2} & \cellcolor{softyellow}\dd{46}{1} & \cellcolor{softyellow} 49.0\\
		DP3~\cite{Ze2024DP3} & \cellcolor{softorange}\dd{83}{1} & \cellcolor{softorange}\dd{63}{2} & \cellcolor{softorange}\dd{82}{4} & \cellcolor{softorange}\dd{46}{2} & \cellcolor{softorange} 68.5\\
		\textbf{DP4 (ours)} & \cellcolor{softred}\ddbf{92}{2} & \cellcolor{softred}\ddbf{84}{1} & \cellcolor{softred}\ddbf{90}{1} & \cellcolor{softred}\ddbf{64}{2}
        & \cellcolor{softred} \textbf{82.5}\\
		\bottomrule
	\end{tabular}
    \vspace{-0.10in}
	\caption{\textbf{Success rates (\%) on the DexArt simulation.} We compare our \textbf{DP4} with more baselines in the DexArt simulation. The baselines include Implicit Behavioral Cloning (IBC), BCRNN, Diffusion Policy (DP), and 3D Diffusion Policy (DP3).}
	\label{table:sim_dexart}
    \vspace{-0.10in}
\end{table}

\noindent \textbf{RLBench simulation}\quad
We compare DP4 with previous state-of-the-art methods on the RLBench task suite. Table~\ref{table:rlbench} presents a comparison of average success rates across tasks. DP4 achieves state-of-the-art performance with average success rates of 63.3\% and 39.9\%, significantly outperforming prior methods, including both perception-based and generative approaches. Leveraging supervision from 3D geometry, texture, and 4D spatiotemporal data, DP4 accurately predicts optimal continuous trajectories from a single RGB-D observation of the current scene.

\begin{table}[t]
\centering
\setlength{\tabcolsep}{0.3pt}
\scriptsize
\begin{tabular}{lccccc|c}
  \toprule
  \textbf{Method} & \makecell{\textit{close} \\ 
  \textit{jar}} & \makecell{\textit{open} \\ \textit{drawer}} & \makecell{\textit{sweep to} \\ \textit{dustpan}} & \makecell{\textit{meat off} \\ \textit{grill}} & \makecell{\textit{turn} \\ \textit{tap}} & \textbf{Overall}\\
  
    \midrule
  PerAct~\cite{shridhar2023perceiver} & \dd{18.7}{8.2} & \dd{54.7}{18.6} & \dd{0.0}{0.0} & \dd{40.0}{17.0} & \dd{38.7}{6.8} & 30.4 \\
  GNFactor~\cite{ze2023gnfactor} & \cellcolor{softyellow}\dd{25.3}{6.8} & \cellcolor{softyellow}\dd{76.0}{5.7} & \cellcolor{softyellow}\dd{28.0}{15.0} & \cellcolor{softyellow}\dd{57.3}{18.9} & \cellcolor{softyellow}\dd{50.7}{8.2} & \cellcolor{softyellow}47.5\\
  ManiGaussian~\cite{lu2024manigaussian} & \cellcolor{softorange}\dd{28.4}{5.4} & \cellcolor{softorange}\dd{76.3}{6.2} & \cellcolor{softorange}\dd{64.5}{13.4} & \cellcolor{softorange}\dd{60.2}{18.2} & \cellcolor{softorange}\dd{56.2}{6.8} & \cellcolor{softorange}57.1 \\
  \textbf{DP4 (ours)} & \cellcolor{softred}\ddbf{35.4}{2.2} & \cellcolor{softred}\ddbf{82.2}{3.4} & \cellcolor{softred}\ddbf{72.0}{5.2} & \cellcolor{softred}\ddbf{64.4}{6.8} & \cellcolor{softred}\ddbf{62.4}{2.8} & \cellcolor{softred}\textbf{63.3} \\
\end{tabular}
\begin{tabular}{lccccc|c}
 \midrule
  \textbf{Method} & \makecell{\textit{slide} \\ \textit{block}} & \makecell{\textit{put in} \\ \textit{drawer}} & \makecell{\textit{drag} \\ \textit{stick}} & \makecell{\textit{push} \\ \textit{buttons}} & \makecell{\textit{stack} \\ \textit{blocks}} & \textbf{Overall}\\
  
\midrule

  PerAct~\cite{shridhar2023perceiver} & \dd{18.7}{13.6} & \dd{2.7}{3.3} & \dd{5.3}{5.0} & \dd{18.7}{12.4} & \dd{6.7}{1.9} & 10.4 \\
  GNFactor~\cite{ze2023gnfactor} & \cellcolor{softyellow}\dd{20.0}{15.0} & \cellcolor{softyellow}\dd{0.0}{0.0} & \cellcolor{softyellow}\dd{37.3}{13.2} & \cellcolor{softyellow}\dd{18.7}{10.0} & \cellcolor{softyellow}\dd{4.0}{3.3} & \cellcolor{softyellow}16.0 \\
  ManiGaussian~\cite{lu2024manigaussian} & \cellcolor{softorange}\dd{24.3}{12.8} & \cellcolor{softorange}\dd{16.6}{3.2} & \cellcolor{softorange}\dd{92.3}{11.4} & \cellcolor{softorange}\dd{20.4}{12.2} & \cellcolor{softorange}\dd{12.4}{3.2} & \cellcolor{softorange}33.2 \\
  \textbf{DP4 (ours)} & \cellcolor{softred}\ddbf{33.4}{4.4} & \cellcolor{softred}\ddbf{22.8}{1.2} & \cellcolor{softred}\ddbf{94.4}{6.2} & \cellcolor{softred}\ddbf{28.3}{6.8} & \cellcolor{softred}\ddbf{21.0}{2.2} & \cellcolor{softred}\textbf{39.9} \\

\bottomrule
\end{tabular}
\vspace{-0.10in}
	\caption{\textbf{Success rates (\%) on the RLBench simulation.} We compare our \textbf{DP4} with more baselines in the RLBench simulation. Baseline approaches include PerAct, GNFactor, and ManiGaussian.}
	\label{table:rlbench}
\vspace{-0.20in}
\end{table}

\subsection{Qualitative Comparison}
\begin{figure}
	\centering	\includegraphics[width=1.0\linewidth]{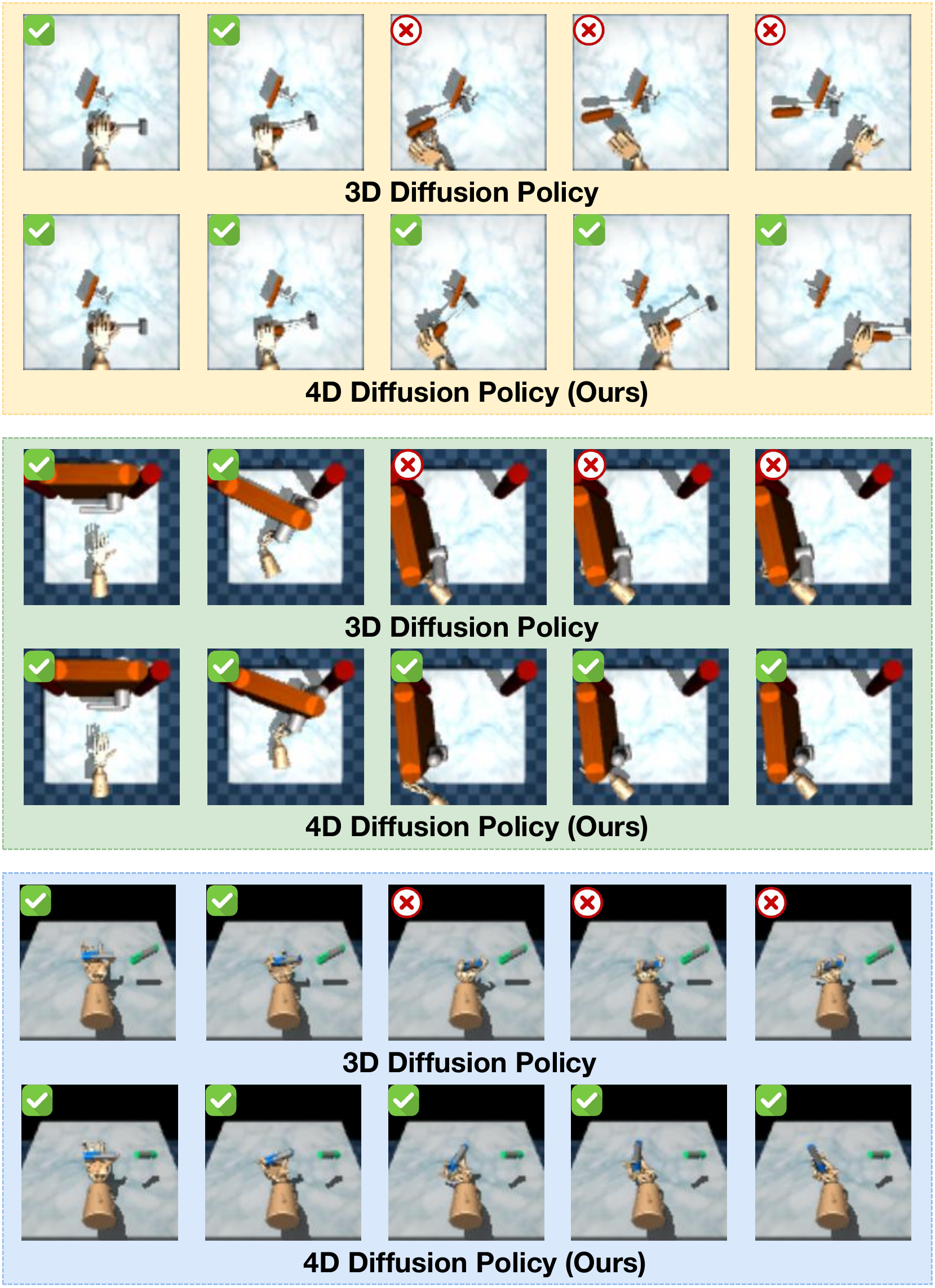}
    \vspace{-0.20in}
	\caption{\textbf{Qualitative case study.} The red mark indicates a pose that significantly deviates from the expert demonstration, while the green mark denotes a pose that aligns with the expert trajectory. The \textbf{4D Diffusion Policy (DP4)} integrates 3D spatial and 4D spatiotemporal awareness with diffusion policies, successfully completing the tasks.
    \label{fig:qual}}
    \vspace{-0.15in}
    \end{figure}

\noindent \textbf{Visualization of  trajectories}\quad We present three qualitative examples of action sequences generated by DP3 and our DP4 in Figure \ref{fig:qual}. The results show that the previous agent (DP3) frequently fails to complete tasks, as it imitates the expert's motions (e.g., hammering nails, sliding doors, and spinning pens) without fully understanding the scene. In contrast, DP4 significantly improves the success rate by accurately interpreting 3D scene structures (e.g., nail positions) and 4D dynamics (e.g., pen rotation).

\begin{figure}
	\centering	\includegraphics[width=1.0\linewidth]{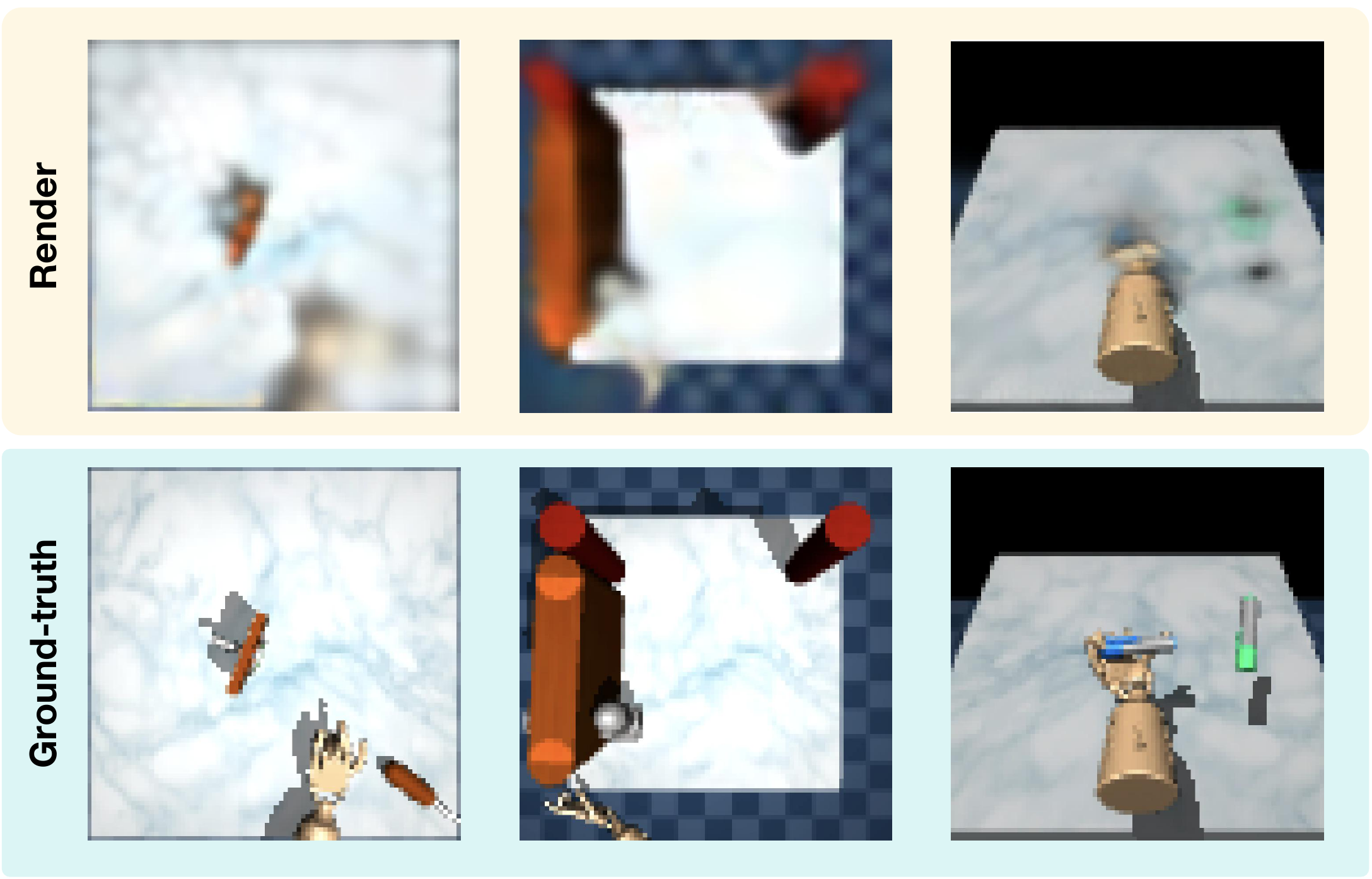}
    \vspace{-0.20in}
	\caption{\textbf{Visualization of Gaussian world model.} After training \textbf{DP4}, we visualize its Gaussian world model by rendering a single-view observation. DP4 uses a Gaussian world model to provide supervision, facilitating the learning of 3D spatial perception.	
    \label{fig:qual_2}}
    \vspace{-0.15in}
\end{figure}

\begin{figure}[t]
    \centering	    \includegraphics[width=0.98\linewidth]{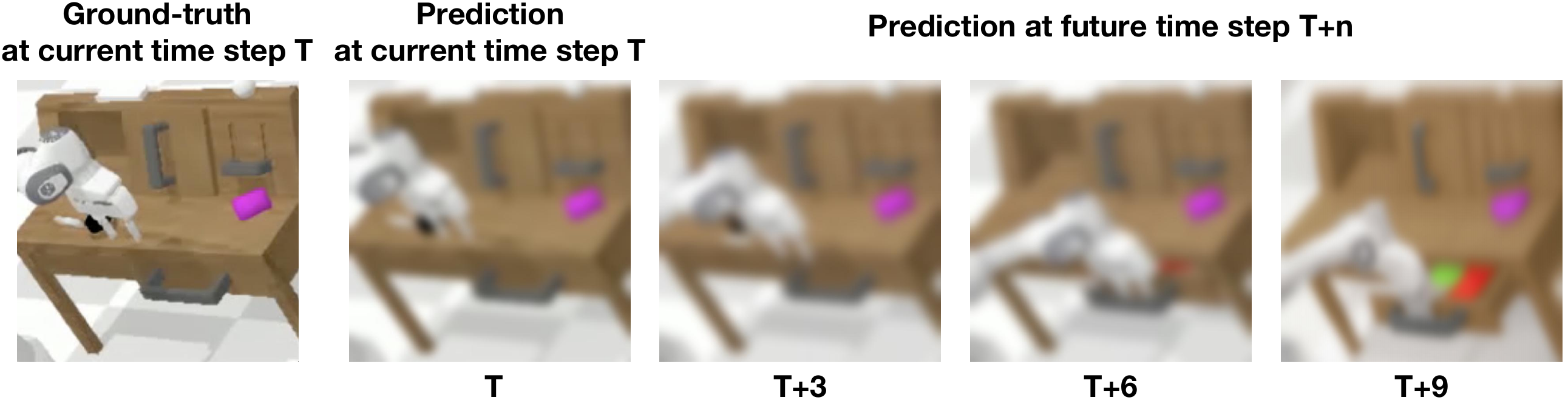}
    \vspace{-0.12in}
	\caption{\textbf{More visualizations of the learned world model.} DP4 uses a Gaussian world model to provide supervision, facilitating the 4D spatiotemporal perceptions in interactive environments.
    \label{fig:qual_3}}
    \vspace{-0.12in}
\end{figure}

\noindent \textbf{Visualization of Gaussian world model}\quad Figure \ref{fig:qual_2} and Figure \ref{fig:qual_3} show the visualization results. We visualize the Gaussian world model via single-view rendering, observing it captures fundamental spatial structures and spatiotemporal dynamics. Although DP4 operates under a realistic single-view constraint, resulting renderings lack fine detail, they effectively represent scene-level 3D structures and 4D dynamics. Crucially, such details have minimal impact on task success; for example, the nail’s thread is irrelevant to the hammering task.

\subsection{Ablation Study} \label{sec.act}
\noindent \textbf{Component ablation.} Ablations are conducted on the Adroit simulation tasks in Table \ref{table:ablation_adr}. Without the proposed components, we use 3D representation without any 3D spatial supervision (RGB and depth (structure)) and 4D spatiotemporal supervision (dynamics). In the experiment on the Adroit simulation, the performance of the ``\textit{Hammer}", ``\textit{Door}", and ``\textit{Pen}" tasks demonstrated improvements in success rates with the proposed components, while the inference time remained stable or only slightly increased.

\begin{itemize}    
    \item \textbf{For the \textit{Hammer} task}: The base model started with a 94.0\% success rate and a completion time of 6.40 seconds. After adding the Gaussian world model with RGB supervision, success improved to 96.0\%, with a minimal increase in time (6.52 seconds). Introducing 3D spatial loss (RGB and Depth supervision) raised success to 95.0\%, with stable time (6.53 seconds). With 4D spatiotemporal supervision (DP4), success reached 98.0\%, and the time increased slightly to 6.56 seconds, indicating a clear performance gain with a small time increase.
    
    \item \textbf{For the \textit{Door} task}: Initially, the model achieved a 94.0\% success rate and a completion time of 6.40 seconds. Adding the Gaussian world model with RGB supervision improved success to 96.0\%, with no significant time change. Incorporating 3D spatial loss (RGB and Depth) increased success to 98.0\%, with a slight increase in time (6.56 seconds). With 4D spatiotemporal supervision (DP4), the success rate reached 100.0\%, with a minimal time increase (6.57 seconds), reflecting significant improvement from additional supervision.

    \item \textbf{For the \textit{Pen} task}: The base model had a 45.0\% success rate and a completion time of 6.46 seconds. After adding the Gaussian world model with RGB supervision, success rose to 48.0\%, with time remaining at 6.48 seconds. Introducing 3D spatial loss (RGB and depth) led to a slight decrease in success (47.0\%). With 4D spatiotemporal supervision (DP4), success jumped to 75.0\%, with a small increase in time to 6.50 seconds, highlighting the effectiveness of 4D supervision.
\end{itemize}

Notably, DP4 does not substantially increase model inference time. This is because DP4 does not construct a Gaussian world model during inference. The 3D spatial and 4D spatiotemporal awareness from the Gaussian world model are efficiently integrated into the multi-level 3D representation during training.

\begin{figure}[t]
	\centering
	\includegraphics[width=\linewidth]{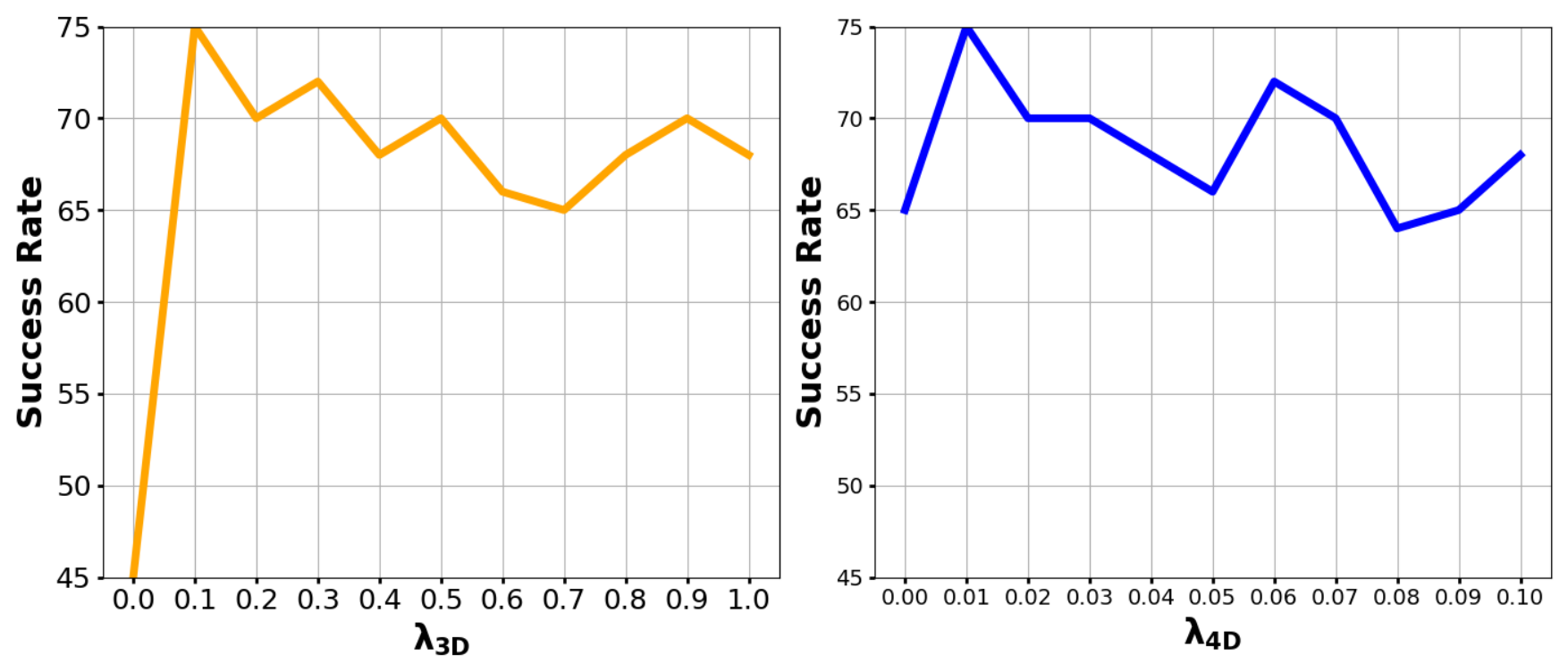}
    \vspace{-0.35in}
	\caption{\textbf{Hyperparameter analysis of DP4.} The effects of $\lambda_{\text{3D}}$ and $\lambda_{\text{4D}}$ on success rates, respectively.
} \label{fig:hyper}	
\vspace{-0.10in}
\end{figure}

\noindent \textbf{Hyperparameter analysis}\quad We conduct hyperparameter analysis on the \textit{Pen} task within the Adroit simulation. DP4 uses two key hyperparameters, $\lambda_{\text{3D}}$ and $\lambda_{\text{4D}}$, to balance the 3D spatial loss for current scenes and the 4D spatiotemporal loss for future scenes across multiple objectives. As shown in Figure \ref{fig:hyper}, we analyze the hyperparameters of DP4. We vary $\lambda_{\text{3D}}$ from 0 to 1 and $\lambda_{\text{4D}}$ from 0 to 0.1, measuring the corresponding success rates. We find that the success rate peaks when $\lambda_{\text{3D}} = 0.1$ and $\lambda_{\text{4D}} = 0.01$, so we adopt these values as the hyperparameters.

\begin{table}[t] \small
	\renewcommand\tabcolsep{1.4pt} 
	\centering
	\begin{tabular}{cc|c|ccc}
		\toprule
		\multicolumn{6}{c}{\textbf{Results on Adroit Simulation}} \\
		\midrule
		\multicolumn{2}{c|}{3D Component} &\multicolumn{1}{c|}{4D Component} & \multicolumn{3}{c}{Performance} \\
		\midrule
		RGB & Depth & Dynamics & \textit{Hammer} & \textit{Door} & \textit{Pen} \\
		\midrule
        &  &  & {94.0/6.40} & {64.0/6.44} & {45.0/6.46} \\
		\CheckmarkBold &  &  & {96.0/6.52} & {68.0/6.45} & {48.0/6.48} \\
        & \CheckmarkBold &  & {95.0/6.53} & {68.0/6.46} & {47.0/6.48} \\
		\CheckmarkBold & \CheckmarkBold &  & {98.0/6.56} & {75.0/6.51} & {72.0/6.47} \\
		\CheckmarkBold & \CheckmarkBold & \CheckmarkBold & {100.0/6.57} & {80.0/6.45} & {75.0/6.50} \\
		\bottomrule
	\end{tabular}	
\vspace{-0.10in}
    \caption{\textbf{Component ablation studies.} We report results for the Adroit simulation tasks. In the performance metrics, the number to the left of $/$ represents the success rate (\%), and the number to the right denotes the inference speed (seconds per task).}
	\label{table:ablation_adr}
    \vspace{-0.18in}
\end{table}

\begin{table}[t] \small
	\setlength{\tabcolsep}{7.2pt} 
	\centering
	\small 
	\begin{tabular}{lcccc}
		\toprule
		\textbf{Method} & 
        \makecell{\textit{Grasping} \\   \textit{Bottles}} & 
        \makecell{\textit{Stacking} \\   \textit{Cups}} &
        \makecell{\textit{Pouring} \\   \textit{Water}} &  \textbf{Overall}  \\
		\midrule
		DP~\cite{chi2023diffusion} & \cellcolor{softyellow}36.0 & \cellcolor{softyellow}44.0 & \cellcolor{softyellow}28.0 & \cellcolor{softyellow}36.0\\
		DP3~\cite{Ze2024DP3} & \cellcolor{softorange}42.0 & \cellcolor{softorange}62.0 & \cellcolor{softorange}34.0 & \cellcolor{softorange}46.0\\
		\textbf{DP4 (ours)} & \cellcolor{softred}\textbf{48.0} & \cellcolor{softred}\textbf{72.0} & \cellcolor{softred}\textbf{44.0}
        & \cellcolor{softred} \textbf{54.6}\\
		\bottomrule
	\end{tabular}
    \vspace{-0.10in}
	\caption{\textbf{Success rates (\%) on the real-robot tasks.} We compare our \textbf{DP4} with additional baselines on the real-robot tasks. We include the Diffusion Policy (DP) and 3D Diffusion Policy (DP3).}
	\label{table:real_world}
    \vspace{-0.25in}
\end{table}

\begin{figure}
	\centering	\includegraphics[width=1.0\linewidth]{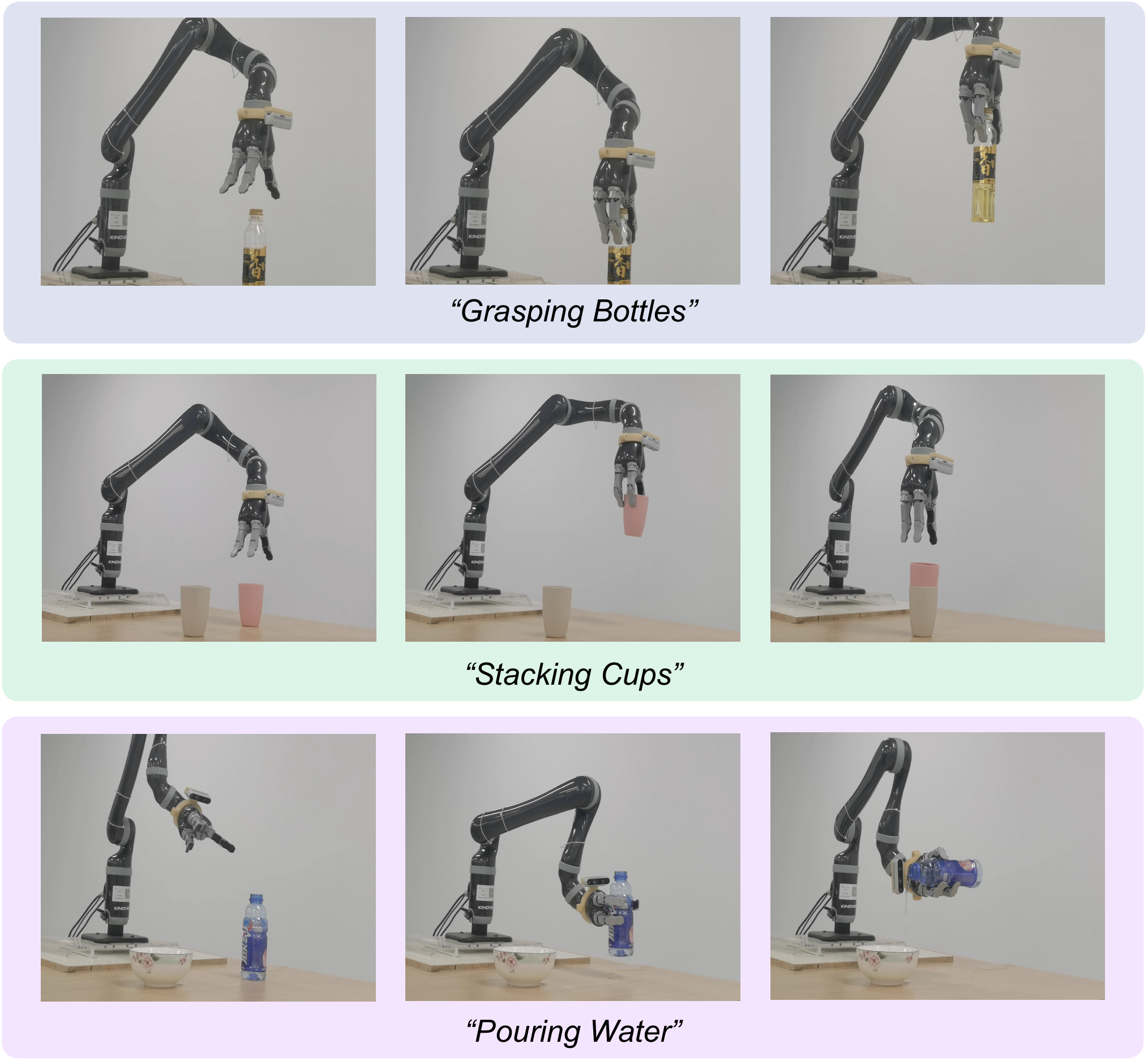}
    \vspace{-0.20in}
	\caption{\textbf{Visualization of DP4 performance on three real-world robotic tasks.}	DP4 demonstrates strong performance in real-world settings and effectively handles a variety of common tasks with a single view. \label{fig:real_vis}}
    \vspace{-0.20in}
\end{figure}

\subsection{Real-Robot Analysis}
To assess DP4's performance on real-world robotic tasks, we designed three tasks: \textit{Grasping Bottles}, \textit{Stacking Cups}, and \textit{Pouring Water}. We focus on the model's performance at key frames, where success is defined as when the error is below a certain threshold. The results of the real-world robotic experiments are summarized in Table \ref{table:real_world}. Experimental results show that DP4 outperforms the Diffusion Policy (DP) and 3D Diffusion Policy (DP3) baselines across all tasks. These results confirm that DP4 effectively integrates spatiotemporal awareness with diffusion policies, improving the accuracy of generated trajectories. Additionally, visualizations of DP4's performance on real-world tasks are shown in Figure \ref{fig:real_vis}.

\section{Conclusion}
This paper presents 4D Diffusion Policy (DP4), which integrates spatiotemporal awareness of the physical world with diffusion-based visual imitation learning. Utilizing a single-view RGB-D input, DP4 constructs a dynamic Gaussian world model that provides rich supervision for acquiring 3D spatial and 4D spatiotemporal perceptions from interactive environments. Extensive experiments on simulated and real-robot tasks validate the effectiveness and superiority.

\section{Acknowledgments}
This work was supported in part by the Science and Technology Commission of Shanghai Municipality (No.24511103100), NSFC Project (62176061),  Doubao Fund, and Shanghai Technology Development and Entrepreneurship Platform for Neuromorphic and AI SoC.

{
    \small
    \bibliographystyle{ieeenat_fullname}
    \bibliography{main}

\begin{thebibliography}{46}
\providecommand{\natexlab}[1]{#1}
\providecommand{\url}[1]{\texttt{#1}}
\expandafter\ifx\csname urlstyle\endcsname\relax
  \providecommand{\doi}[1]{doi: #1}\else
  \providecommand{\doi}{doi: \begingroup \urlstyle{rm}\Url}\fi

\bibitem[Arunachalam et~al.(2023)Arunachalam, Silwal, Evans, and Pinto]{arunachalam2023dexterous}
Sridhar~Pandian Arunachalam, Sneha Silwal, Ben Evans, and Lerrel Pinto.
\newblock Dexterous imitation made easy: A learning-based framework for efficient dexterous manipulation.
\newblock In \emph{2023 ieee international conference on robotics and automation (icra)}, pages 5954--5961. IEEE, 2023.

\bibitem[Bao et~al.(2023)Bao, Xu, Qin, and Wang]{bao2023dexart}
Chen Bao, Helin Xu, Yuzhe Qin, and Xiaolong Wang.
\newblock Dexart: Benchmarking generalizable dexterous manipulation with articulated objects.
\newblock In \emph{Proceedings of the IEEE/CVF Conference on Computer Vision and Pattern Recognition}, pages 21190--21200, 2023.

\bibitem[Chen et~al.(2021)Chen, Lu, Rajeswaran, Lee, Grover, Laskin, Abbeel, Srinivas, and Mordatch]{chen2021decision}
Lili Chen, Kevin Lu, Aravind Rajeswaran, Kimin Lee, Aditya Grover, Misha Laskin, Pieter Abbeel, Aravind Srinivas, and Igor Mordatch.
\newblock Decision transformer: Reinforcement learning via sequence modeling.
\newblock \emph{Advances in neural information processing systems}, 34:\penalty0 15084--15097, 2021.

\bibitem[Chen et~al.(2023)Chen, Sun, Song, and Luo]{chen2023diffusiondet}
Shoufa Chen, Peize Sun, Yibing Song, and Ping Luo.
\newblock Diffusiondet: Diffusion model for object detection.
\newblock In \emph{Proceedings of the IEEE/CVF international conference on computer vision}, pages 19830--19843, 2023.

\bibitem[Chi et~al.(2023)Chi, Xu, Feng, Cousineau, Du, Burchfiel, Tedrake, and Song]{chi2023diffusion}
Cheng Chi, Zhenjia Xu, Siyuan Feng, Eric Cousineau, Yilun Du, Benjamin Burchfiel, Russ Tedrake, and Shuran Song.
\newblock Diffusion policy: Visuomotor policy learning via action diffusion.
\newblock \emph{The International Journal of Robotics Research}, page 02783649241273668, 2023.

\bibitem[Du et~al.(2022)Du, Ho, Alemi, Jang, and Khansari]{du2022bayesian}
Yuqing Du, Daniel Ho, Alex Alemi, Eric Jang, and Mohi Khansari.
\newblock Bayesian imitation learning for end-to-end mobile manipulation.
\newblock In \emph{International Conference on Machine Learning}, pages 5531--5546. PMLR, 2022.

\bibitem[Florence et~al.(2022)Florence, Lynch, Zeng, Ramirez, Wahid, Downs, Wong, Lee, Mordatch, and Tompson]{florence2022implicit}
Pete Florence, Corey Lynch, Andy Zeng, Oscar~A Ramirez, Ayzaan Wahid, Laura Downs, Adrian Wong, Johnny Lee, Igor Mordatch, and Jonathan Tompson.
\newblock Implicit behavioral cloning.
\newblock In \emph{Conference on robot learning}, pages 158--168. PMLR, 2022.

\bibitem[Ha and Schmidhuber(2018)]{ha2018recurrent}
David Ha and J{\"u}rgen Schmidhuber.
\newblock Recurrent world models facilitate policy evolution.
\newblock \emph{Advances in neural information processing systems}, 31, 2018.

\bibitem[Hafner et~al.(2019)Hafner, Lillicrap, Ba, and Norouzi]{hafner2019dream}
Danijar Hafner, Timothy Lillicrap, Jimmy Ba, and Mohammad Norouzi.
\newblock Dream to control: Learning behaviors by latent imagination.
\newblock \emph{arXiv preprint arXiv:1912.01603}, 2019.

\bibitem[Hafner et~al.(2020)Hafner, Lillicrap, Norouzi, and Ba]{hafner2020mastering}
Danijar Hafner, Timothy Lillicrap, Mohammad Norouzi, and Jimmy Ba.
\newblock Mastering atari with discrete world models.
\newblock \emph{arXiv preprint arXiv:2010.02193}, 2020.

\bibitem[Hafner et~al.(2022)Hafner, Lee, Fischer, and Abbeel]{hafner2022deep}
Danijar Hafner, Kuang-Huei Lee, Ian Fischer, and Pieter Abbeel.
\newblock Deep hierarchical planning from pixels.
\newblock \emph{Advances in Neural Information Processing Systems}, 35:\penalty0 26091--26104, 2022.

\bibitem[Haldar et~al.(2023)Haldar, Pari, Rai, and Pinto]{haldar2023teach}
Siddhant Haldar, Jyothish Pari, Anant Rai, and Lerrel Pinto.
\newblock Teach a robot to fish: Versatile imitation from one minute of demonstrations.
\newblock \emph{arXiv preprint arXiv:2303.01497}, 2023.

\bibitem[Hansen et~al.(2022)Hansen, Yuan, Ze, Mu, Rajeswaran, Su, Xu, and Wang]{hansen2022pre}
Nicklas Hansen, Zhecheng Yuan, Yanjie Ze, Tongzhou Mu, Aravind Rajeswaran, Hao Su, Huazhe Xu, and Xiaolong Wang.
\newblock On pre-training for visuo-motor control: Revisiting a learning-from-scratch baseline.
\newblock \emph{arXiv preprint arXiv:2212.05749}, 2022.

\bibitem[Ho et~al.(2022)Ho, Salimans, Gritsenko, Chan, Norouzi, and Fleet]{ho2022video}
Jonathan Ho, Tim Salimans, Alexey Gritsenko, William Chan, Mohammad Norouzi, and David~J Fleet.
\newblock Video diffusion models.
\newblock \emph{Advances in Neural Information Processing Systems}, 35:\penalty0 8633--8646, 2022.

\bibitem[Hu et~al.(2022)Hu, Corrado, Griffiths, Murez, Gurau, Yeo, Kendall, Cipolla, and Shotton]{hu2022model}
Anthony Hu, Gianluca Corrado, Nicolas Griffiths, Zachary Murez, Corina Gurau, Hudson Yeo, Alex Kendall, Roberto Cipolla, and Jamie Shotton.
\newblock Model-based imitation learning for urban driving.
\newblock \emph{Advances in Neural Information Processing Systems}, 35:\penalty0 20703--20716, 2022.

\bibitem[Hu et~al.(2023)Hu, Russell, Yeo, Murez, Fedoseev, Kendall, Shotton, and Corrado]{hu2023gaia}
Anthony Hu, Lloyd Russell, Hudson Yeo, Zak Murez, George Fedoseev, Alex Kendall, Jamie Shotton, and Gianluca Corrado.
\newblock Gaia-1: A generative world model for autonomous driving.
\newblock \emph{arXiv preprint arXiv:2309.17080}, 2023.

\bibitem[James et~al.(2020)James, Ma, Arrojo, and Davison]{james2020rlbench}
Stephen James, Zicong Ma, David~Rovick Arrojo, and Andrew~J Davison.
\newblock Rlbench: The robot learning benchmark \& learning environment.
\newblock \emph{IEEE Robotics and Automation Letters}, 5\penalty0 (2):\penalty0 3019--3026, 2020.

\bibitem[Johns(2021)]{johns2021coarse}
Edward Johns.
\newblock Coarse-to-fine imitation learning: Robot manipulation from a single demonstration.
\newblock In \emph{2021 IEEE international conference on robotics and automation (ICRA)}, pages 4613--4619. IEEE, 2021.

\bibitem[Kerbl et~al.(2023)Kerbl, Kopanas, Leimk{\"u}hler, and Drettakis]{kerbl20233d}
Bernhard Kerbl, Georgios Kopanas, Thomas Leimk{\"u}hler, and George Drettakis.
\newblock 3d gaussian splatting for real-time radiance field rendering.
\newblock \emph{ACM Trans. Graph.}, 42\penalty0 (4):\penalty0 139--1, 2023.

\bibitem[Kim et~al.(2024)Kim, Pertsch, Karamcheti, Xiao, Balakrishna, Nair, Rafailov, Foster, Lam, Sanketi, Vuong, Kollar, Burchfiel, Tedrake, Sadigh, Levine, Liang, and Finn]{kim24openvla}
{Moo Jin} Kim, Karl Pertsch, Siddharth Karamcheti, Ted Xiao, Ashwin Balakrishna, Suraj Nair, Rafael Rafailov, Ethan Foster, Grace Lam, Pannag Sanketi, Quan Vuong, Thomas Kollar, Benjamin Burchfiel, Russ Tedrake, Dorsa Sadigh, Sergey Levine, Percy Liang, and Chelsea Finn.
\newblock Openvla: An open-source vision-language-action model.
\newblock \emph{arXiv preprint arXiv:2406.09246}, 2024.

\bibitem[Li et~al.(2025)Li, Wu, Huang, Cheang, Wang, and Kong]{li2025gr}
Peiyan Li, Hongtao Wu, Yan Huang, Chilam Cheang, Liang Wang, and Tao Kong.
\newblock Gr-mg: Leveraging partially-annotated data via multi-modal goal-conditioned policy.
\newblock \emph{IEEE Robotics and Automation Letters}, 2025.

\bibitem[Liu et~al.(2024)Liu, Wu, Li, Tan, Chen, Wang, Xu, Su, and Zhu]{liu2024rdt}
Songming Liu, Lingxuan Wu, Bangguo Li, Hengkai Tan, Huayu Chen, Zhengyi Wang, Ke Xu, Hang Su, and Jun Zhu.
\newblock Rdt-1b: a diffusion foundation model for bimanual manipulation.
\newblock \emph{arXiv preprint arXiv:2410.07864}, 2024.

\bibitem[Lu et~al.(2024)Lu, Zhang, Wang, Liu, Lu, and Tang]{lu2024manigaussian}
Guanxing Lu, Shiyi Zhang, Ziwei Wang, Changliu Liu, Jiwen Lu, and Yansong Tang.
\newblock Manigaussian: Dynamic gaussian splatting for multi-task robotic manipulation.
\newblock In \emph{European Conference on Computer Vision}, pages 349--366. Springer, 2024.

\bibitem[Mandlekar et~al.(2021)Mandlekar, Xu, Wong, Nasiriany, Wang, Kulkarni, Fei-Fei, Savarese, Zhu, and Mart{\'\i}n-Mart{\'\i}n]{mandlekar2021matters}
Ajay Mandlekar, Danfei Xu, Josiah Wong, Soroush Nasiriany, Chen Wang, Rohun Kulkarni, Li Fei-Fei, Silvio Savarese, Yuke Zhu, and Roberto Mart{\'\i}n-Mart{\'\i}n.
\newblock What matters in learning from offline human demonstrations for robot manipulation.
\newblock \emph{arXiv preprint arXiv:2108.03298}, 2021.

\bibitem[Mao et~al.(2024)Mao, Xu, Zhan, Zhang, and Zhang]{mao2024diffusion}
Liyuan Mao, Haoran Xu, Xianyuan Zhan, Weinan Zhang, and Amy Zhang.
\newblock Diffusion-dice: In-sample diffusion guidance for offline reinforcement learning.
\newblock \emph{arXiv preprint arXiv:2407.20109}, 2024.

\bibitem[Peng et~al.(2020)Peng, Coumans, Zhang, Lee, Tan, and Levine]{peng2020learning}
Xue~Bin Peng, Erwin Coumans, Tingnan Zhang, Tsang-Wei Lee, Jie Tan, and Sergey Levine.
\newblock Learning agile robotic locomotion skills by imitating animals.
\newblock \emph{arXiv preprint arXiv:2004.00784}, 2020.

\bibitem[Qin et~al.(2022)Qin, Wu, Liu, Jiang, Yang, Fu, and Wang]{qin2022dexmv}
Yuzhe Qin, Yueh-Hua Wu, Shaowei Liu, Hanwen Jiang, Ruihan Yang, Yang Fu, and Xiaolong Wang.
\newblock Dexmv: Imitation learning for dexterous manipulation from human videos.
\newblock In \emph{European Conference on Computer Vision}, pages 570--587. Springer, 2022.

\bibitem[Rajeswaran et~al.(2017)Rajeswaran, Kumar, Gupta, Vezzani, Schulman, Todorov, and Levine]{rajeswaran2017learning}
Aravind Rajeswaran, Vikash Kumar, Abhishek Gupta, Giulia Vezzani, John Schulman, Emanuel Todorov, and Sergey Levine.
\newblock Learning complex dexterous manipulation with deep reinforcement learning and demonstrations.
\newblock \emph{arXiv preprint arXiv:1709.10087}, 2017.

\bibitem[Schulman et~al.(2017)Schulman, Wolski, Dhariwal, Radford, and Klimov]{schulman2017proximal}
John Schulman, Filip Wolski, Prafulla Dhariwal, Alec Radford, and Oleg Klimov.
\newblock Proximal policy optimization algorithms.
\newblock \emph{arXiv preprint arXiv:1707.06347}, 2017.

\bibitem[Seo et~al.(2023)Seo, Hafner, Liu, Liu, James, Lee, and Abbeel]{seo2023masked}
Younggyo Seo, Danijar Hafner, Hao Liu, Fangchen Liu, Stephen James, Kimin Lee, and Pieter Abbeel.
\newblock Masked world models for visual control.
\newblock In \emph{Conference on Robot Learning}, pages 1332--1344. PMLR, 2023.

\bibitem[Shafiullah et~al.(2022)Shafiullah, Cui, Altanzaya, and Pinto]{shafiullah2022behavior}
Nur~Muhammad Shafiullah, Zichen Cui, Ariuntuya~Arty Altanzaya, and Lerrel Pinto.
\newblock Behavior transformers: Cloning $ k $ modes with one stone.
\newblock \emph{Advances in neural information processing systems}, 35:\penalty0 22955--22968, 2022.

\bibitem[Shafiullah et~al.(2023)Shafiullah, Rai, Etukuru, Liu, Misra, Chintala, and Pinto]{shafiullah2023bringing}
Nur Muhammad~Mahi Shafiullah, Anant Rai, Haritheja Etukuru, Yiqian Liu, Ishan Misra, Soumith Chintala, and Lerrel Pinto.
\newblock On bringing robots home.
\newblock \emph{arXiv preprint arXiv:2311.16098}, 2023.

\bibitem[Shridhar et~al.(2023)Shridhar, Manuelli, and Fox]{shridhar2023perceiver}
Mohit Shridhar, Lucas Manuelli, and Dieter Fox.
\newblock Perceiver-actor: A multi-task transformer for robotic manipulation.
\newblock In \emph{Conference on Robot Learning}, pages 785--799. PMLR, 2023.

\bibitem[Song et~al.(2020)Song, Meng, and Ermon]{song2020denoising}
Jiaming Song, Chenlin Meng, and Stefano Ermon.
\newblock Denoising diffusion implicit models.
\newblock \emph{arXiv:2010.02502}, 2020.

\bibitem[Wang et~al.(2022{\natexlab{a}})Wang, Luo, Ross, and Li]{wang2022vrl3}
Che Wang, Xufang Luo, Keith Ross, and Dongsheng Li.
\newblock Vrl3: A data-driven framework for visual deep reinforcement learning.
\newblock \emph{Advances in Neural Information Processing Systems}, 35:\penalty0 32974--32988, 2022{\natexlab{a}}.

\bibitem[Wang et~al.(2023)Wang, Fan, Sun, Zhang, Fei-Fei, Xu, Zhu, and Anandkumar]{wang2023mimicplay}
Chen Wang, Linxi Fan, Jiankai Sun, Ruohan Zhang, Li Fei-Fei, Danfei Xu, Yuke Zhu, and Anima Anandkumar.
\newblock Mimicplay: Long-horizon imitation learning by watching human play.
\newblock \emph{arXiv preprint arXiv:2302.12422}, 2023.

\bibitem[Wang et~al.(2024)Wang, Zhu, Huang, Chen, Zhu, and Lu]{wang2024drivedreamer}
Xiaofeng Wang, Zheng Zhu, Guan Huang, Xinze Chen, Jiagang Zhu, and Jiwen Lu.
\newblock Drivedreamer: Towards real-world-drive world models for autonomous driving.
\newblock In \emph{European Conference on Computer Vision}, pages 55--72. Springer, 2024.

\bibitem[Wang et~al.(2022{\natexlab{b}})Wang, Hunt, and Zhou]{wang2022diffusion}
Zhendong Wang, Jonathan~J Hunt, and Mingyuan Zhou.
\newblock Diffusion policies as an expressive policy class for offline reinforcement learning.
\newblock \emph{arXiv preprint arXiv:2208.06193}, 2022{\natexlab{b}}.

\bibitem[Wu et~al.(2024)Wu, Yi, Fang, Xie, Zhang, Wei, Liu, Tian, and Wang]{wu20244d}
Guanjun Wu, Taoran Yi, Jiemin Fang, Lingxi Xie, Xiaopeng Zhang, Wei Wei, Wenyu Liu, Qi Tian, and Xinggang Wang.
\newblock 4d gaussian splatting for real-time dynamic scene rendering.
\newblock In \emph{Proceedings of the IEEE/CVF conference on computer vision and pattern recognition}, pages 20310--20320, 2024.

\bibitem[Wu et~al.(2023)Wu, Escontrela, Hafner, Abbeel, and Goldberg]{wu2023daydreamer}
Philipp Wu, Alejandro Escontrela, Danijar Hafner, Pieter Abbeel, and Ken Goldberg.
\newblock Daydreamer: World models for physical robot learning.
\newblock In \emph{Conference on robot learning}, pages 2226--2240. PMLR, 2023.

\bibitem[Yang et~al.(2023{\natexlab{a}})Yang, Huang, Lei, Zhong, Yang, Fang, Wen, Zhou, and Lin]{yang2023policy}
Long Yang, Zhixiong Huang, Fenghao Lei, Yucun Zhong, Yiming Yang, Cong Fang, Shiting Wen, Binbin Zhou, and Zhouchen Lin.
\newblock Policy representation via diffusion probability model for reinforcement learning.
\newblock \emph{arXiv preprint arXiv:2305.13122}, 2023{\natexlab{a}}.

\bibitem[Yang et~al.(2023{\natexlab{b}})Yang, Zhang, Song, Hong, Xu, Zhao, Zhang, Cui, and Yang]{yang2023diffusion}
Ling Yang, Zhilong Zhang, Yang Song, Shenda Hong, Runsheng Xu, Yue Zhao, Wentao Zhang, Bin Cui, and Ming-Hsuan Yang.
\newblock Diffusion models: A comprehensive survey of methods and applications.
\newblock \emph{ACM Computing Surveys}, 56\penalty0 (4):\penalty0 1--39, 2023{\natexlab{b}}.

\bibitem[Yang et~al.(2023{\natexlab{c}})Yang, Chen, Ma, Zheng, Chen, Nguyen, and Wang]{yang2023generalized}
Ruihan Yang, Zhuoqun Chen, Jianhan Ma, Chongyi Zheng, Yiyu Chen, Quan Nguyen, and Xiaolong Wang.
\newblock Generalized animal imitator: Agile locomotion with versatile motion prior.
\newblock \emph{arXiv preprint arXiv:2310.01408}, 2023{\natexlab{c}}.

\bibitem[Ze et~al.(2023)Ze, Yan, Wu, Macaluso, Ge, Ye, Hansen, Li, and Wang]{ze2023gnfactor}
Yanjie Ze, Ge Yan, Yueh-Hua Wu, Annabella Macaluso, Yuying Ge, Jianglong Ye, Nicklas Hansen, Li~Erran Li, and Xiaolong Wang.
\newblock Gnfactor: Multi-task real robot learning with generalizable neural feature fields.
\newblock In \emph{Conference on Robot Learning}, pages 284--301. PMLR, 2023.

\bibitem[Ze et~al.(2024)Ze, Zhang, Zhang, Hu, Wang, and Xu]{Ze2024DP3}
Yanjie Ze, Gu Zhang, Kangning Zhang, Chenyuan Hu, Muhan Wang, and Huazhe Xu.
\newblock 3d diffusion policy: Generalizable visuomotor policy learning via simple 3d representations.
\newblock In \emph{Proceedings of Robotics: Science and Systems (RSS)}, 2024.

\bibitem[Zhu et~al.(2025)Zhu, Wang, Huang, Ye, Ouyang, and He]{zhu2025point}
Haoyi Zhu, Yating Wang, Di Huang, Weicai Ye, Wanli Ouyang, and Tong He.
\newblock Point cloud matters: Rethinking the impact of different observation spaces on robot learning.
\newblock \emph{Advances in Neural Information Processing Systems}, 37:\penalty0 77799--77830, 2025.

\end{thebibliography}
}

\clearpage
\setcounter{page}{1}
\maketitlesupplementary
\setcounter{section}{0}

\section{Video}
The attached video demonstrates the application of our 4D Diffusion Policy, which is capable of handling complex tasks in real-world through high-level  spatial-temporal awareness. 
We present an expert demonstration and evaluate the performance of DP4 after learning from this demonstration to complete similar tasks in other environments, thereby demonstrating the generalization ability of DP4.	
The following is a detailed explanation of the tasks that the 4D Diffusion Policy handles in the video:

\noindent \textbf{Short-term Task: "Grasping Bottles"} In the first scenario, the 4D Diffusion Policy demonstrates its ability to accurately grasp a bottle by leveraging its sophisticated 3D spatial awareness. This task involves precise control and interaction with a bottle placed within a dynamic environment. The policy processes sensory input and calculates the correct approach trajectory, ensuring the gripper is aligned with the bottle's position and orientation. Through this task, the 4D Diffusion Policy showcases its short-term task efficiency, where it must focus on accurately detecting and physically interacting with the object.
\begin{itemize}
    \item The policy achieves this by continually adjusting its movements based on real-time feedback and predict the bottle's location, ensuring successful manipulation.
    \item The success of this task highlights the policy's ability to integrate real-time spatial information into a coherent action plan, achieving reliable object manipulation.
\end{itemize}

\noindent \textbf{Long-term Task: ``Stacking Cups"} The second scenario involves a more complex, long-term task—stacking cups. The 4D Diffusion Policy exhibits its ability to grasp a cup and stack it onto another cup, all while understanding the high-level scene structures. This task requires not only precise interaction with the cups but also an awareness of the scene's dynamic configuration. The policy must predict and plan for the interaction between multiple objects (i.e., the cups) and their positions in a broader context.
\begin{itemize}
    \item The 4D Diffusion Policy relies on a Gaussian world model to map the environment's layout, enabling it to recognize and process object placements and potential changes in the scene's configuration. The Gaussian world model allows for the modeling of uncertainties and complex interactions between objects, enhancing the policy's adaptability and accuracy in high-level scene tasks.
    \item In this case, the policy anticipates the need to stack the cups while considering factors like object stability and gravity, ensuring that the final configuration is stable.
\end{itemize}

\noindent \textbf{Fluid Dynamics Task: ``Pouring Water"} The final task demonstrates the 4D Diffusion Policy's ability to handle complex fluid dynamics. In this scenario, the policy must interact with both a liquid (water) and a container, demonstrating its advanced understanding of fluid behavior and high-level scene dynamics. This task involves pouring water from one container to another while maintaining a steady stream and avoiding spills.
\begin{itemize}
    \item The policy takes into account the fluid's interaction with the container and its movement through space. By understanding the physics of fluid dynamics, the 4D Diffusion Policy can anticipate the water's behavior and adjust its actions accordingly to successfully complete the task.
    \item The use of fluid dynamics modeling, along with high-level scene understanding, allows the policy to adjust its grasping strength and pouring angle, ensuring that the water is transferred without splashing or spillage.
\end{itemize}

\section{Additional Implementation Details}
The proposed 4D Diffusion Policy (DP4) comprises three key components: Multi-level 3D Spatial Awareness, 4D Spatiotemporal Awareness, and Diffusion-based Decision. Below, we provide a detailed description of the implementation of each component.	

\noindent\textbf{Multi-level 3D Spatial Awareness.} To effectively capture complex 3D structures in the interactive physical world, this component primarily includes the 3D Local Encoder, 3D Global Encoder, and Generalizable Gaussian Regressor.	
\begin{itemize}
    \item The input to the 3D Local Encoder consists of a colorless cropped point cloud, downsampled from the raw point cloud using Farthest Point Sampling (FPS). We use a sample size of 512 in both simulated and real-world tasks. DP4 encodes the cropped point cloud into a compact 3D local representation using our designed DP4 3D Local Encoder. Below, we present a simple PyTorch implementation of the DP4 3D Local Encoder.	
\begin{lstlisting}[language=Python, frame=none, basicstyle=\small\ttfamily, commentstyle=\color{ourblue}\small\ttfamily,columns=fullflexible, breaklines=true, postbreak=\mbox{\textcolor{red}{$\hookrightarrow$}\space}, escapeinside={(*}{*)}]
class DP4LocalEncoder(nn.Module):
    def __init__(self, 
         observation_space: Dict, 
         out_channel=256,
         state_mlp_size=(64, 64),
         voxel_mlp_size=(32, 32),
         use_pc_color=False,
         pointnet_type='pointnet'):
        super().__init__()
        
        self.n_output_channels = out_channel
        self.point_cloud_shape = observation_space['point_cloud']
        self.state_shape = observation_space['agent_pos']
        
        self.use_pc_color = use_pc_color
        self.pointnet_type = pointnet_type
        
        if pointnet_type == "pointnet":
            self.extractor = PointNetEncoderXYZRGB() if use_pc_color else PointNetEncoderXYZ()

        # State MLP
        self.state_mlp = nn.Sequential(MLP(self.state_shape[0], state_mlp_size[-1], state_mlp_size[:-1], nn.ReLU))

        self.n_output_channels += state_mlp_size[-1]
\end{lstlisting}
    \item To comprehensively capture global 3D information, the 3D Global Encoder receives input in the form of the global voxel derived from the full point cloud and the robot's pose. DP4 processes this voxel to produce a compact global 3D representation through our custom-designed DP4 3D Global Encoder. A simple PyTorch implementation of the DP4 3D Global Encoder is presented:	
\begin{lstlisting}[language=Python, frame=none, basicstyle=\small\ttfamily, commentstyle=\color{ourblue}\small\ttfamily,columns=fullflexible, breaklines=true, postbreak=\mbox{\textcolor{red}{$\hookrightarrow$}\space}, escapeinside={(*}{*)}]
class DP4GlobalEncoder(nn.Module):
    def __init__(self, in_channels=10, out_channels=64, norm_act=InPlaceABN):
        super().__init__()
        CHANNELS = [8, 16, 32, 64]
        # Define the convolution layers
        self.conv0 = ConvBnReLU3D(in_channels, CHANNELS[0], norm_act=norm_act)
        self.conv1 = ConvBnReLU3D(CHANNELS[0], CHANNELS[1], stride=2, norm_act=norm_act)
        self.conv2 = ConvBnReLU3D(CHANNELS[1], CHANNELS[1], norm_act=norm_act)
        self.conv3 = ConvBnReLU3D(CHANNELS[1], CHANNELS[2], stride=2, norm_act=norm_act)
        self.conv4 = ConvBnReLU3D(CHANNELS[2], CHANNELS[2], norm_act=norm_act)
        self.conv5 = ConvBnReLU3D(CHANNELS[2], CHANNELS[3], stride=2, norm_act=norm_act)
        self.conv6 = ConvBnReLU3D(CHANNELS[3], CHANNELS[3], norm_act=norm_act)
        # Transpose convolutions for upsampling
        self.conv7 = nn.Sequential(
            nn.ConvTranspose3d(CHANNELS[3], CHANNELS[2], 3, padding=1, stride=2, bias=False),
            norm_act(CHANNELS[2]))
        self.conv9 = nn.Sequential(
            nn.ConvTranspose3d(CHANNELS[2], CHANNELS[1], 3, padding=1, output_padding=1, stride=2, bias=False),
            norm_act(CHANNELS[1])
        )
        self.conv11 = nn.Sequential(
            nn.ConvTranspose3d(CHANNELS[1], CHANNELS[0], 3, padding=1, output_padding=1, stride=2, bias=False),
            norm_act(CHANNELS[0]) )

        # Final convolution for output
        self.conv_out = nn.Conv3d(CHANNELS[0], out_channels, 1, stride=1, padding=0, bias=True)

\end{lstlisting}
    \item To improve the representation with global structural and texture data, we employ a Generalizable Gaussian Regressor. This regressor uses the deep volume as a scene representation to derive the Gaussian world model via Gaussian Splatting. The model is trained by rendering both RGB and depth images based on the generated Gaussian world model. Below, we present a simple PyTorch implementation of Generalizable Gaussian Regressor.	
\begin{lstlisting}[language=Python, frame=none, basicstyle=\small\ttfamily, commentstyle=\color{ourblue}\small\ttfamily,columns=fullflexible, breaklines=true, postbreak=\mbox{\textcolor{red}{$\hookrightarrow$}\space}, escapeinside={(*}{*)}]
class GeneralizableGSRegressor(nn.Module):
    def __init__(self, cfg, with_gs_render=True):
        super().__init__()
        self.cfg = cfg
        self.with_gs_render = with_gs_render

        self.use_xyz = cfg.use_xyz
        self.d_in = 3 if self.use_xyz else 1
        self.use_code = cfg.use_code

        if self.use_code:
            self.code = PositionalEncoding.from_conf(cfg["code"], d_in=self.d_in)
            self.d_in = self.code.d_out

        self.d_latent = cfg.d_latent
        self.d_lang = cfg.d_lang
        self.d_out = sum(self._get_splits_and_inits(cfg)[0])

        self.encoder = ResnetFC(
            d_in=self.d_in, d_latent=self.d_latent, d_lang=self.d_lang, d_out=self.d_out,
            d_hidden=cfg.mlp.d_hidden, n_blocks=cfg.mlp.n_blocks, combine_layer=cfg.mlp.combine_layer,
            beta=cfg.mlp.beta, use_spade=cfg.mlp.use_spade,
        )
        self.gs_parm_regresser = GSPointCloudRegresser(cfg, self._get_splits_and_inits(cfg)[0])

        self.scaling_activation = torch.exp
        self.opacity_activation = torch.sigmoid
        self.rotation_activation = torch.nn.functional.normalize

\end{lstlisting}

\end{itemize}

\noindent\textbf{4D Spatiotemporal Awareness.} To highlight 4D spatiotemporal awareness within the diffusion policy, our DP4 incorporates dynamics into the Gaussian world model. Expanding on this model, DP4 utilizes a deformable MLP to track changes in Gaussian parameters over time. Below, we present a simple PyTorch implementation of the DP4 deformable MLP.	
\begin{lstlisting}[language=Python, frame=none, basicstyle=\small\ttfamily, commentstyle=\color{ourblue}\small\ttfamily,columns=fullflexible, breaklines=true, postbreak=\mbox{\textcolor{red}{$\hookrightarrow$}\space}, escapeinside={(*}{*)}]
class DeformableNet(nn.Module):
    def __init__(self, d_in, d_out=4, n_blocks=5, d_latent=0, d_hidden=128, beta=0.0, combine_layer=1000, use_spade=False):
        super().__init__()

        self.lin_in = nn.Linear(d_in, d_hidden) if d_in > 0 else None
        self.lin_out = nn.Linear(d_hidden, d_out)
        self.blocks = nn.ModuleList([ResnetBlockFC(d_hidden, beta=beta) for _ in range(n_blocks)])

        self.d_latent = d_latent
        self.use_spade = use_spade
        self.combine_layer = combine_layer

        # Initialize the layers
        if self.lin_in:
            nn.init.kaiming_normal_(self.lin_in.weight, a=0, mode="fan_in")
            nn.init.constant_(self.lin_in.bias, 0.0)
        nn.init.kaiming_normal_(self.lin_out.weight, a=0, mode="fan_in")
        nn.init.constant_(self.lin_out.bias, 0.0)

        # Handle latent layers
        self.lin_z = nn.ModuleList([nn.Linear(d_latent, d_hidden) for _ in range(min(combine_layer, n_blocks))])
        self.scale_z = nn.ModuleList([nn.Linear(d_latent, d_hidden) for _ in range(min(combine_layer, n_blocks))]) if use_spade 

        # Activation function
        self.activation = nn.Softplus(beta=beta) if beta > 0 else nn.ReLU()

\end{lstlisting}

\noindent\textbf{Diffusion-based Decision.} The Diffusion-based Decision backbone is a convolutional network-based diffusion policy that converts random Gaussian noise into a coherent sequence of actions. For implementation, we employ the official PyTorch framework provided by 3D Diffusion Policy (DP3)~\cite{Ze2024DP3}.	In practice, the model is designed to predict a series of $H$ actions based on $N_{obs}$ observed timesteps, but it executes only the last $N_{act}$ actions during inference.	We set $H=4$, $N_{obs}=2$, and $N_{act}=3$ for the diffusion-based baselines, which are consistent with the original DP3 configuration.	We independently scale the minimum and maximum values of each action and observation dimension to $[-1, 1]$. Normalizing actions to $[-1, 1]$ is essential for DDPM and DDIM predictions, as these models clip the predictions to $[-1, 1]$ for stability.		

\section{Details of Simulation Tasks}
For the simulation experiments, we selected tasks from various domains, covering a broad spectrum of robotic skills.	These tasks include both challenging scenarios, such as bimanual manipulation, deformable object manipulation, and articulated object manipulation, and simpler tasks like parallel gripper manipulation.	The tasks consist of Adroit, DexArt, and RLBench tasks.	This section provides a detailed overview of these simulation tasks.	

\noindent \textbf{Adroit.} This task employs a simulated version of the highly dexterous manipulator ADROIT, a 24-DoF anthropomorphic platform designed to tackle challenges in dynamic and precise manipulation. The first, middle, and ring fingers have 4 DoF, while the little finger and thumb have 5 DoF, and the wrist has 2 DoF. Each DoF is controlled by position control and is equipped with a joint angle sensor. The experimental setup of ADROIT utilizes the MuJoCo physics simulator. MuJoCo's stable contact dynamics make it well-suited for contact-rich hand manipulation tasks.	The kinematics, dynamics, and sensing details of the physical hardware were meticulously modeled to enhance physical realism.	
In addition to dry friction in the joints, all hand-object interactions involve planar friction.	
Object-fingertip interactions support torsion and rolling friction.	

\noindent \textbf{DexArt.} This simulation task includes four dexterous manipulation tasks: Faucet, Bucket, Laptop, and Toilet, each involving a mix of seen and unseen objects. The Faucet task evaluates the coordination between the dexterous hand and arm motions. Although a 2-jaw parallel gripper could perform this task, it heavily depends on precise arm motion due to its low DoF end-effector. In the Bucket task, the robot must lift a bucket. To ensure stability, it should extend its hand under the handle and grip it to achieve form closure. In the Laptop task, the robot grasps the middle of the screen and opens the laptop lid, a task well-suited for dexterous hands. A parallel gripper can achieve this by precisely gripping the lid between its jaws. The Toilet task is similar to the Laptop task, but requires the robot to open a larger, more irregular toilet lid. This task is more challenging due to the lid's irregular and diverse geometry.
\begin{figure}
	\centering	\includegraphics[width=1.0\linewidth]{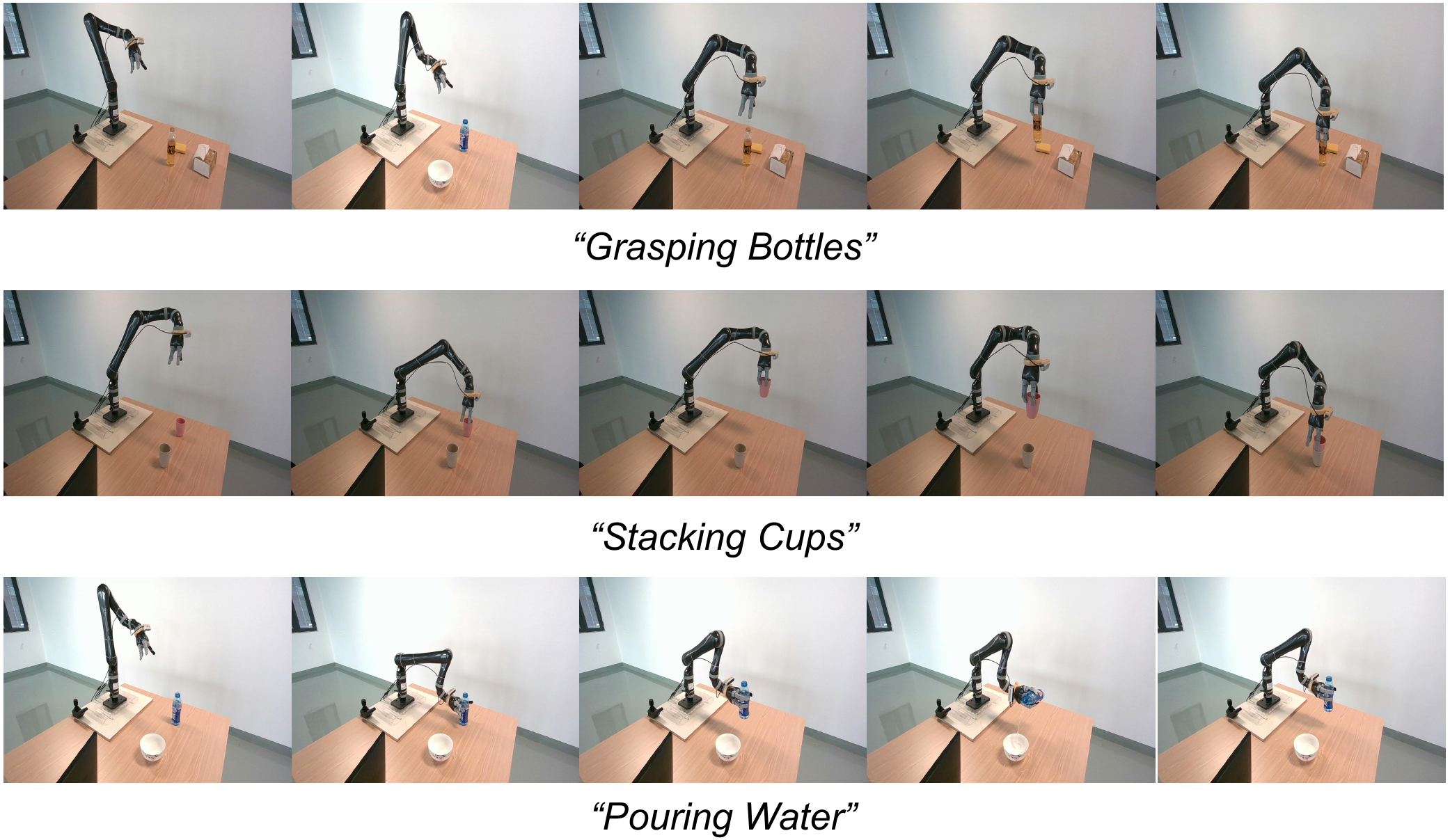}
    \vspace{-0.20in}
	\caption{\textbf{Keyframes for real robot tasks.} We give the keyframes used in our 3 real robot tasks.
    \label{fig:supp}}
    \vspace{-0.20in}
    \end{figure}
    
\noindent \textbf{RLBench.}
We selected 10 tasks from RLBench~\cite{james2020rlbench}, each with a minimum of two variations. These variations include random changes in color, size, count, placement, and object category, resulting in 166 unique combinations. The set of colors includes 20 options: red, maroon, lime, green, blue, navy, yellow, cyan, magenta, silver, gray, orange, olive, purple, teal, azure, violet, rose, black, and white. The size variations are categorized as short and tall, while the count options are 1, 2, and 3. The placement and object categories differ depending on the task. For instance, in the "open drawer" task, objects can be placed in one of three locations: top, middle, or bottom. Additionally, objects are randomly positioned on the tabletop within a constrained pose range. We also created six extra tasks where the scene is modified from the original training environment to assess the system's generalization capability.

\section{Details of Real-Robot Tasks}
In the experiments, we conduct three primary tasks, along with three additional ones that include distracting objects. The ``\textit{grasping bottles}" task requires the agent to pick up a bottle from a table, a challenge due to the precise coordination and 3D spatial awareness needed. The ``\textit{pouring water}" task involves transferring water from a bottle to a bowl, requiring the agent to manage the complex 4D dynamics of the water in the physical world. The ``\textit{stacking cups}" task tasks the agent with finding a randomly placed cup on the table and understanding its 3D spatial structure. Among these tasks, the ``\textit{pouring water}" task is considered the most difficult, as it demands accurate placement and rotation of the gripper in response to the positioning. The keyframes for the real robot tasks are shown in Figure \ref{fig:supp}.

\end{document}